\documentclass{article}

\usepackage{amssymb}
\usepackage{enumitem}
\usepackage{comment}
\usepackage[pagebackref,breaklinks,colorlinks]{hyperref}
\usepackage{xspace}
\usepackage{pifont}
\usepackage{verbatim}
\usepackage{tcolorbox}
\usepackage{xcolor}
\usepackage{float}
\usepackage{multicol}
\usepackage{listings}
\usepackage{pythonhighlight}
\usepackage{tikz}
\usepackage{amsmath}
\usepackage{cleveref}
\usepackage{booktabs}
\usepackage{wrapfig}

\definecolor{keywordcolor}{RGB}{0,0,255} 
\definecolor{commentcolor}{RGB}{34,139,34} 
\definecolor{stringcolor}{RGB}{178,34,34} 
\definecolor{backgroundcolor}{RGB}{245,245,245} 
\definecolor{argcolor}{RGB}{128,0,0} 
\definecolor{functioncolor}{RGB}{128,0,128} 

\usepackage{tcolorbox}
\tcbuselibrary{skins, breakable}

\tcbset{
    mymonobox/.style={
        enhanced,
        colback=black!5,               
        colframe=black!80,             
        sharp corners=southwest,      
        fonttitle=\bfseries,          
        coltitle=black,               
        breakable,
        drop shadow=black!50,         
        boxrule=0.5mm,                
        before skip=10pt,             
        after skip=10pt,              
        left=2mm,                     
        right=2mm,                    
        top=2mm,                      
        bottom=2mm,                   
        before upper={\ttfamily\footnotesize},      
        width=\textwidth             
    }
}

\definecolor{boxcolor}{RGB}{37, 150, 190} 

\newcommand{\ourmethod}{GraphEQA\xspace}

\newcommand{\para}[1]{\noindent\textbf{#1}}

\usepackage{multirow}
\usepackage{wrapfig}

\usepackage[final]{corl_2025} 

\usepackage{titlesec}

\titlespacing{\section}{0pt}{*1}{*0.25}
\titlespacing{\subsection}{0pt}{*1}{*0.10}
\titlespacing{\subsubsection}{0pt}{*1}{*0.10}

\title{\ourmethod: Using 3D Semantic Scene Graphs \\ for Real-time Embodied Question Answering}
\author{Saumya Saxena$^{1}$\thanks{Equal contribution. \textsuperscript{\ding{41}}Correspondence.}~~\textsuperscript{\ding{41}}
\And Blake Buchanan$^{2}$\footnotemark[1]~  \enspace Chris Paxton$^{3}$ \enspace Peiqi Liu$^{4}$ \enspace Bingqing Chen$^{5}$
\And Narunas Vaskevicius$^{5}$ \enspace Luigi Palmieri$^{5}$ \enspace Jonathan Francis$^{1,5}$ \enspace Oliver Kroemer$^{1}$
\\
\\
{$^{1}$Carnegie Mellon University\enspace
$^{2}$Neya Systems\enspace
$^{3}$Agility Robotics\enspace
$^{4}$Hello Robot\enspace
$^{5}$Bosch Center for AI}
\\
{\tt\small \{saumyas,okroemer\}@andrew.cmu.edu, bbuchanan@neyarobotics.com, chris.paxton.cs@gmail.com}, \\
{\tt\small pl2285@nyu.edu},
{\tt\small \{bingqing.chen,jon.francis\}@us.bosch.com}, \\
{\tt\small \{narunas.vaskevicius,luigi.palmieri\}@de.bosch.com}
\thanks{This work was in part supported by the National Science Foundation under Grant No. CMMI-1925130 and in part by the EU Horizon 2020 research and innovation program under grant agreement No. 101017274 (DARKO). Any opinions, findings, and conclusions or recommendations expressed in this material are those of the author(s) and do not necessarily reflect the views of the NSF.}
}

\begin{document}
\maketitle
\vspace{-2.5em}
\begin{abstract} %
In Embodied Question Answering (EQA), agents must explore and develop a semantic understanding of an unseen environment to answer a situated question with confidence. This problem remains challenging in robotics, due to the difficulties in obtaining useful semantic representations, updating these representations online, and leveraging prior world knowledge for efficient planning and exploration. To address these limitations, we propose \ourmethod, a novel approach that utilizes real-time 3D metric-semantic scene graphs (3DSGs) and task relevant images as multi-modal memory for grounding Vision-Language Models (VLMs) to perform EQA tasks in unseen environments. We employ a hierarchical planning approach that exploits the hierarchical nature of 3DSGs for structured planning and semantics-guided exploration. We evaluate \ourmethod in simulation on two benchmark datasets, HM-EQA and OpenEQA, and demonstrate that it outperforms key baselines by completing EQA tasks with higher success rates and fewer planning steps. We further demonstrate \ourmethod in multiple real-world home and office environments. Videos and code: \href{https://saumyasaxena.github.io/grapheqa/}{website}.





\end{abstract}

\section{Introduction}
\label{sec:intro}


Embodied Question Answering (EQA) \cite{das2018embodied} is a challenging task in robotics, wherein an agent is required to explore and understand a previously unseen environment sufficiently well, to answer an embodied question in natural language. Accomplishing this task efficiently requires agents to rely on both commonsense knowledge of human environments as well as ground its exploration strategy in the current environment context. For example, to answer the question ``How many chairs are there at the dining table?", the agent might rely on commonsense knowledge to understand that dining tables are often associated with dining rooms and dining rooms are usually near the kitchen towards the back of a home. A reasonable navigation strategy would involve navigating to the back of the house to locate a kitchen. To ground this search in the current environment, however, requires the agent to continually maintain an understanding of where it is, memory of where it has been, and what further exploratory actions will lead it to relevant regions. Finally, the agent needs to observe the target object(s) and perform visual grounding to reason about the number of chairs around the dining table, and confidently answer the question correctly.

Maintaining a concise and effective memory and using it to take actions in the environment is critical for solving EQA tasks. Prior works have demonstrated the impressive commonsense reasoning capabilities of Vision Language Models (VLMs) as planning agents, while leveraging a semantic map for retrieval \cite{gu2023conceptgraphsopenvocabulary3dscene} or semantic exploration \cite{ren2024exploreconfident}. In such approaches, the VLMs are not grounded in the current environment, and commonsense reasoning and context-based decision-making are disconnected. Recent works \cite{rana2023sayplan,shah2024bumbleunifyingreasoningacting, anwar2024remembrbuildingreasoninglonghorizon, xie2024embodiedraggeneralnonparametricembodied, xumobility} focus on maintaining memory modules that can be queried by VLM agents for grounded planning. To construct a semantically rich memory, prior works either maintain a large, extensive set of images \cite{xumobility, majumdar2024openeqa} or have to perform an expensive offline processing step to obtain a compact representation \cite{anwar2024remembrbuildingreasoninglonghorizon, xie2024embodiedraggeneralnonparametricembodied, xu2024vlm}.
Thus, such semantic memory modules are either semantically rich \cite{xie2024embodiedraggeneralnonparametricembodied, gu2023conceptgraphsopenvocabulary3dscene, hovsg}, compact \cite{rana2023sayplan, shah2024bumbleunifyingreasoningacting}, \textit{or} online \cite{shah2024bumbleunifyingreasoningacting}, but not all at the same time. 

To address these limitations, we propose \ourmethod, an approach for embodied question answering that builds an online, compact, multimodal semantic memory combining global, semantically-sparse, and task-agnostic information from real-time 3D scene graphs \cite{hughes2022hydra} with local and semantically-rich information from task-relevant images \cite{xu2024vlm}. GraphEQA uses this multimodal representation for grounding vision-language planners to solve EQA tasks in unseen environments.
Specifically, we utilize a recent spatial perception system \cite{hughes2022hydra} that incrementally creates a real-time 3D metric-semantic hierarchical scene graph (3DSG), given sequential egocentric image frames. We further augment this scene graph with semantic room labels and semantically-enriched frontiers, while maintaining a task-relevant visual memory that keeps track of task-relevant images as the robot explores the environment. Finally, we employ a hierarchical planning approach that utilizes the hierarchical nature of scene graphs and semantically relevant frontiers for structured planning and exploration in an unseen environment before using the multimodal memory to answer the embodied question with high confidence.

\begin{figure}[t]
    \centering
\includegraphics[width=0.98\textwidth]{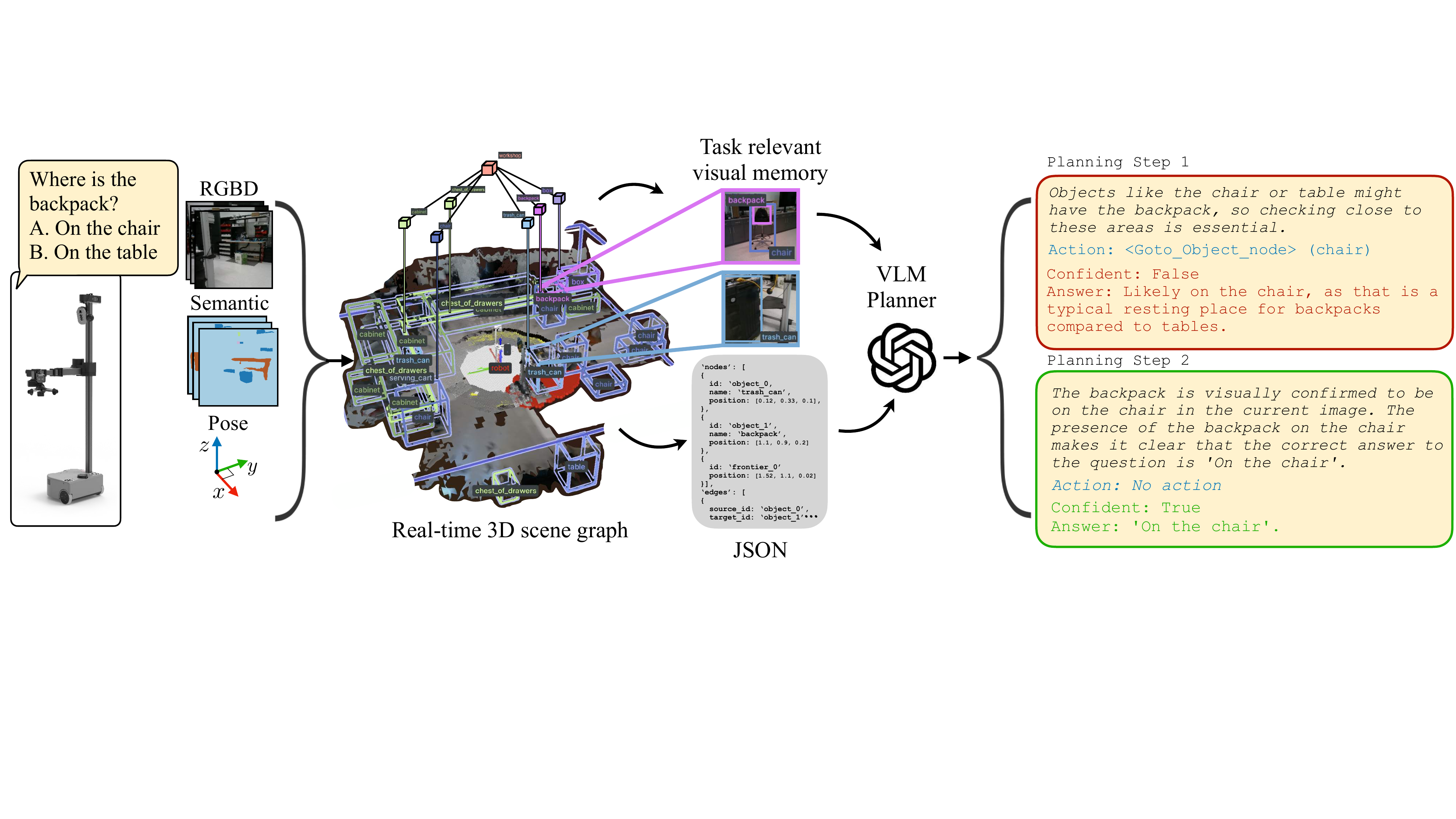}
    \vspace{-7pt}
    \caption{\footnotesize{\textbf{Overview of \ourmethod:} A novel approach for utilizing real-time 3D metric-semantic hierarchical scene graphs and task-relevant images as multimodal memory for grounding vision-language based planners to solve embodied question answering tasks in unseen environments.}}
    \vspace{-18pt}
    \label{fig:teaser}
\end{figure}

We demonstrate that given our real-time multimodal memory and hierarchical planning approach, the agent is able to accomplish long-horizon tasks with significantly fewer VLM planning steps, explores explainable task-relevant frontiers, and succeeds in EQA tasks with a higher rate than previous approaches. We demonstrate our results on the HM-EQA \cite{ren2024exploreconfident} and OpenEQA \cite{OpenEQA2023} datasets in the Habitat simulation environment \cite{yadav2022habitat} and also in the real world using the Hello Robot Stretch mobile manipulator in two scenes.
While the focus of this work is EQA tasks, our contributions advance the broader goal of grounding vision-language foundation models in unseen 3D environments, making GraphEQA applicable to a wider range of robotics tasks.
The key contributions of this work are as follows:%
\vspace{-5pt}
\begin{itemize}[itemsep=0.05cm, parsep=0.05cm]
    \item We present \ourmethod, a novel approach for using real-time 3D metric-semantic hierarchical scene graphs and task-relevant images as multimodal memory, for grounding VLMs to solve EQA tasks in unseen environments.
    \item We propose an approach to enrich 3DSGs with 1) semantically enriched frontiers and 2) semantic room labels.
    \item We propose a hierarchical VLM-based planning approach that exploits the hierarchical nature of the enriched 3DSG for structured exploration and planning.
    \item We provide extensive experimental results, both in simulation on the HM-EQA \cite{ren2024exploreconfident} and OpenEQA \cite{OpenEQA2023} datasets, and in the real-world in two indoor environments, using the Hello Robot Stretch RE2 mobile manipulator.
\end{itemize}

\section{Related Work}
\vspace{-5pt}
\label{sec:relatedwork}
\noindent\textbf{3D Semantic Scene Graph Representations for Planning:}
3D semantic scene graphs (3DSGs) \cite{armeni20193dsg, kim2020, rosinol2021kimeraslam, wald2020learning, greve2024collaborative} have emerged as compact, semantically-rich representations of indoor environments, spurring advances in both offline \cite{hovsg, gu2023conceptgraphsopenvocabulary3dscene, li2024llmenhancedscenegraphlearning, chang2023contextawareentitygroundingopenvocabulary} and online \cite{hughes2022hydra, maggio2024Clio, wu2021scenegraphfusion} prediction. Offline approaches \cite{hovsg, gu2023conceptgraphsopenvocabulary3dscene} focus on enriching 3DSG nodes and edges with open-vocabulary semantics via Vision-Language Models (VLMs) \cite{kirillov2023segment, radford2021learning}, supporting retrieval-based, language-guided tasks. Online methods \cite{hughes2022hydra, maggio2024Clio} enable real-time deployment of embodied agents but often rely on closed-set semantics \cite{hughes2022hydra} or fixed task sets \cite{maggio2024Clio}, limiting open-world generalization. Our approach bridges these paradigms by building a \textit{multimodal memory}: an online-constructed 3DSG with closed-set semantics guides a VLM agent toward task-relevant areas to capture semantically rich images \cite{xu2024vlm}, enabling open-world embodied question answering (EQA). Recent work has leveraged textual 3DSG representations for VLM-based planning \cite{rana2023sayplan, MOMALLM_Honerkamp_2024, rajvanshi2024saynavgroundinglargelanguage, agia2022taskography}, primarily targeting object search or rearrangement tasks, but without addressing the deeper semantic understanding required for EQA. Other spatio-temporal representations of environments have also been effective for planning, e.g., \cite{liu2024dynamemonlinedynamicspatiosemantic}, which constructs a custom 3D data structure to maintain dynamic memory.

\noindent\textbf{VLMs for 3D Scene Understanding and Planning:}
Several previous works leverage the commonsense reasoning capabilities of foundation models for long-horizon planning \cite{ahn2022icanisay, huang2022languagemodelszeroshotplanners, huang2022inner}. However, these methods are not grounded in the context of the current environment and additional tools are required to translate the LLM plan to executable actions \cite{chen2023open, liang2023code, huang2022languagemodelszeroshotplanners, song2023llm}. 
Previous works have explored the use of VLMs for building dense queryable open-vocabulary 3D semantic representations using explicit pixel-level \cite{ding2023pla,peng2023openscene,huang2023visual,zhang2023clip, jatavallabhula2023conceptfusion} or implicit neural \cite{shafiullah2022clip, kerr2023lerf, tschernezki2022neural} representations. However, these maps are built offline, before they are used for downstream retrieval-based tasks. Moreover, such dense representations are not scalable to large environments and cannot be used to ground VLM-based planners.
Recent advancements in grounding VLM-based planners using videos \cite{zhang2024navidvideobasedvlmplans, xumobility} are promising, but struggle with scalability for long-horizon tasks in large environments. VLMs have been used as planners while leveraging semantic maps for retrieval \cite{chang2023goatthing,gu2023conceptgraphsopenvocabulary3dscene, hovsg} or semantic exploration \cite{ren2024exploreconfident, zhou2023esc, yokoyama2024vlfm, shah2023navigation, nasiriany2024pivot}, however such approaches disconnect context-based decision-making and commonsense reasoning. Offline methods that build topological maps \cite{shah2023lm}, keyframe selections \cite{xu2024vlm}, 3D semantic graphs \cite{gu2023conceptgraphsopenvocabulary3dscene, hovsg, qiu2024openvocabularymobilemanipulationunseen, li2024llmenhancedscenegraphlearning, chang2023contextawareentitygroundingopenvocabulary} and experience summaries \cite{anwar2024remembrbuildingreasoninglonghorizon, xie2024embodiedraggeneralnonparametricembodied, bustamante2024grounding, wu2023tidybot} are unsuitable for real-time deployment in novel settings. Online semantic scene graphs, while real-time, are limited by closed-set semantics. Our approach introduces an online, compact, and semantically rich multimodal memory to effectively ground VLM planners for EQA tasks.

\noindent\textbf{Embodied Question Answering:} Embodied Question Answering \cite{das2018embodied, gordon2018iqa, wijmans2019embodied, francis2022core} has emerged as a challenging paradigm for testing robotic task planning systems on their ability to incrementally build a semantic understanding of an environment in order to correctly answer an embodied question with confidence. 
\citet{ren2024exploreconfident} build an explicit task-specific 2D semantic map of the environment to guide exploration, \citet{anwar2024remembrbuildingreasoninglonghorizon, xie2024embodiedraggeneralnonparametricembodied} build offline experience modules that the LLM can query, and \citet{majumdar2024openeqa} uses video memory to answer embodied questions using long-context VLMs. 
We focus on building agents that do not disconnect the semantic memory from the planner by grounding the planner in a compact scene representation for solving EQA tasks online.



\vspace{-5pt}
\section{Method}
\vspace{-5pt}
\subsection{Preliminaries}
\vspace{-5pt}
\label{prelim}
\label{prelim:hydra}
\label{prelim:tsdf}
\noindent\textbf{Hierarchical 3D Metric-semantic Scene Graphs.} 3D metric-semantic scene graphs (3DSGs) provide a structured, layered representation of environments and encode spatial, semantic, and relational information \cite{armeni20193dsg, kim2020, rosinol2021kimeraslam}. Recent works like Clio \cite{maggio2024Clio}, Hydra \cite{hughes2022hydra}, and Open Scene Graphs \cite{loo2024openscenegraphsopen} introduce efficient real-time frameworks for incremental construction of hierarchical SG layers consisting of objects, regions, rooms, buildings, etc. See Appendix \ref{sec:hydra_sgs} for details.

\noindent\textbf{2D Occupancy Mapping and Frontier Detection.} 3D voxel-based occupancy maps are an effective way for storing explored, occupied, and unexplored regions of an environment for planning and navigation. As the robot explores, using depth data and camera intrinsics, occupancy of the voxels is updated using Volumetric Truncated Signed Distance Function (TSDF) fusion. TSDF integrates depth observations to update voxels as occupied or free, while areas beyond a certain threshold are marked unexplored. Typically, the 3D occupancy map is projected into 2D, where \textit{frontiers}---boundaries between explored and unexplored regions---are identified to guide further exploration. We employ this approach in our method for identifying frontiers, clustering them and adding them to the scene graph.

\begin{figure*}[t]
    \centering
\includegraphics[width=0.95\textwidth]{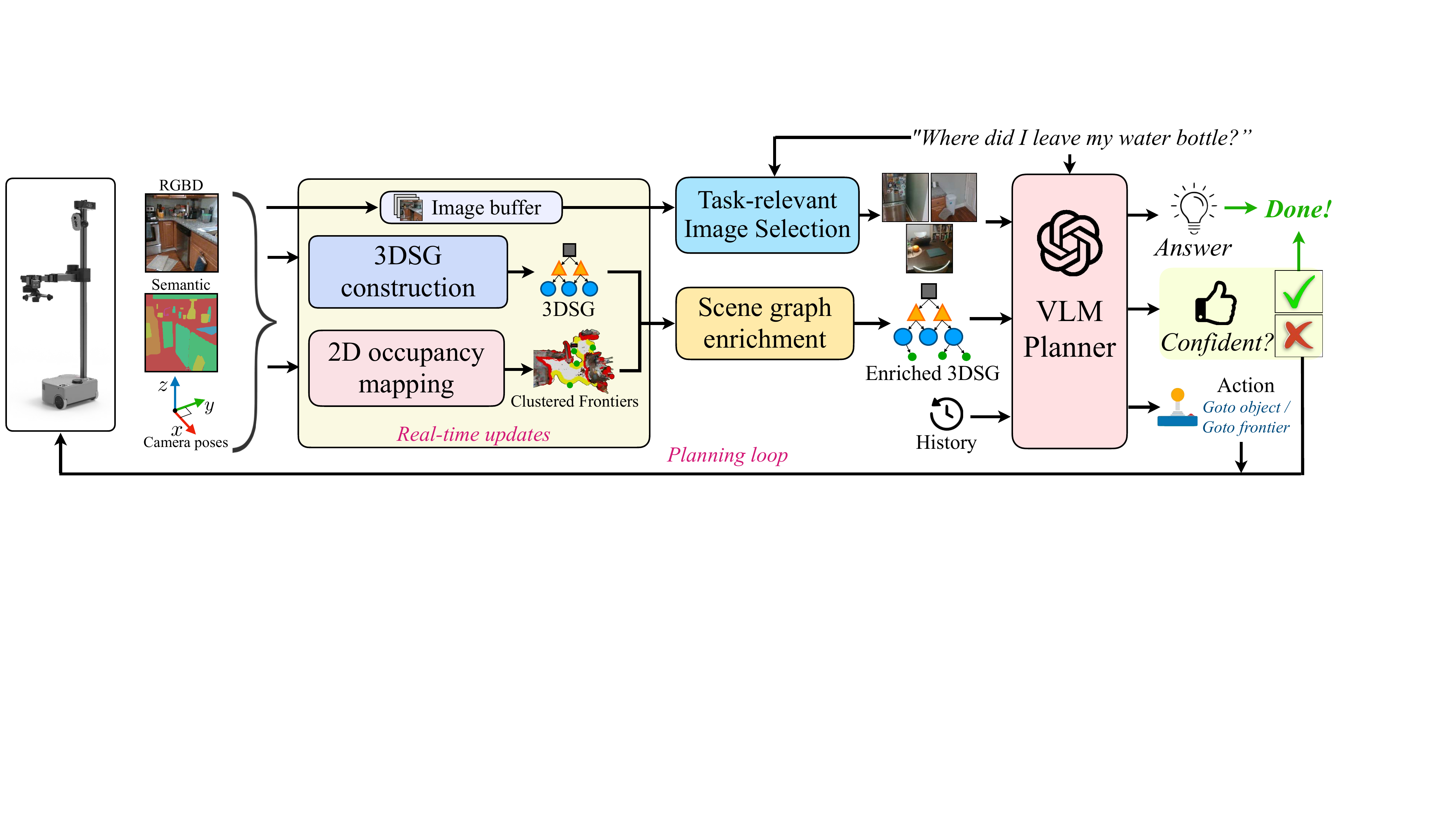}
    \vspace{-5pt}
    \caption{\footnotesize{\textbf{Overall \ourmethod architecture.} As the agent explores the environment, it used its sensor data (RGBD images, semantic map, camera poses and intrinsics) to construct a 3D metric-semantic hierarchical scene graph (3DSG) as well as a 2D occupancy map for frontier selection in real time. The constructed 3DSG is enriched as discussed in~\Cref{enrichment}. From the set of images collected during each trajectory execution, a task-relevant subset is selected, called the task-relevant visual memory (~\Cref{keyframe}). A VLM-based planner (~\Cref{planner}) takes as input the enriched scene graph, task-relevant visual memory, a history of past states and actions, and the embodied question and outputs the answer, its confidence in the selected answer, and the next step it needs to take in the environment. If the VLM agent is confident in its answer, the episode is terminated, else the proposed action is executed in the environment and the process repeats.}}
    \label{fig:method_architecture}
    \vspace{-10pt}
\end{figure*}
\vspace{-5pt}
\subsection{Problem Formulation}
\vspace{-5pt}
An overview of our proposed method is shown in ~\Cref{fig:method_architecture}. Given a multiple choice question $q$, we wish to find the correct answer $a^{*} \in \mathcal{A}$ where $\mathcal{A}$ is the set of multiple-choice answers to question $q$ available to the agent. To find $a^{*}$, the agent is equipped with a VLM-based planner $\mathcal{V}(q, \mathcal{S}^e_{t}, \{\mathcal{I}_k\}^K_{k=1}, \mathcal{H}_{t}, X_t) = (a_{t}, c_{t}, u_{t})$, where $\mathcal{S}^e_{t}$ is the enriched scene graph constructed online at planning time $t$, and includes frontier nodes from 2D the occupancy map (\Cref{prelim:tsdf}). $\{\mathcal{I}_k\}^K_{k=1}$ is a small set of task-relevant images maintained in memory (\Cref{keyframe}), $a_{t}$ is the current best answer to the multiple choice question $q$ and $u_{t}$ is the next action the agent should take in the environment. $\mathcal{H}_{t}$ represents the accrued history up to time ${t}$ and $X_t$ represents the current state of the agent. We query the VLM planner $\mathcal{V}$ (\Cref{planner}) at time $t$ with the inputs described above, with the scene graph $\mathcal{S}^e_{t}$ continually being constructed and a set of images $\{\mathcal{I}_k\}^K_{k=1}$ chosen based on task and semantic relevance. The planner then outputs a high-level action $u_t$ which is executed in the environment while the scene graph, visual memory, and frontiers are all updated in real time. In the following sections we provide details for each of these components.
\vspace{-5pt}
\subsection{Scene Graph Construction and Enrichment}
\vspace{-5pt}
\label{enrichment}
We use Hydra~\cite{hughes2022hydra} to construct a layered metric-semantic scene graph (see Appendix \ref{sec:hydra_sgs}), while also maintaining a 2D occupancy map of the environment depicting the explored, occupied, and unexplored navigable regions of the environment as mentioned in~\Cref{prelim:tsdf}. We perform room and frontier enrichment steps to enable semantic-guided exploration and hierarchical planning.\\
\noindent\textbf{Room enrichment:} Room nodes in Hydra's 3DSG are assigned generic labels such as R0, R1, etc. To enrich them with semantic information, we prompt an LLM to assign semantic labels to each of the room nodes. We use a simple prompt ``Which room are these objects \textless object list\textgreater~most likely located in?'' where \textless object list\textgreater~is the list of all objects located in a certain room in the scene graph. The output of the LLM is then used to update the room labels.\\
\noindent\textbf{Frontier enrichment:} 
To enrich our 3DSG with semantic information that can enable task-relevant exploration, we extract frontier nodes from the 2D occupancy map, cluster them, and add them as independent nodes to the scene graph. Next, we find top-j object nodes nearest to each clustered frontier node, within a maximum distance $d$. We add edges to the scene graph connecting each frontier node to its top-j object neighbors. This semantic information can now be utilized by a VLM-based planner to select the most semantically-relevant frontier to explore next. For general exploration, for example, it could be useful to choose frontier nodes near doors. We use $j=3$ and $d=2$ meters in our experiments, but can be varied based on the environment.
\vspace{-5pt}
\subsection{Task-relevant Visual Memory}
\vspace{-5pt}
\label{keyframe}
During action-execution, images are stored in a buffer at a specified sampling frequency to avoid multiple similar repeated images. Images from this buffer, along with keywords from the question/task, are then processed using SigLIP \cite{zhai2023siglip} to obtain the text-image relevancy score for each image. Using this score, only the top-K most relevant images are maintained in the buffer and the rest are discarded. We use $K=2$; we append these K images together, along with the agent's current view, and use it as the visual input to the VLM planner at the next planning step, as shown in ~\Cref{fig:method_architecture}.
\vspace{-8pt}
\subsection{Hierarchical Vision-Language Planner}
\vspace{-5pt}
\label{planner}
\begin{wrapfigure}{r}{0.5\textwidth}
    \vspace{-20pt}
    \centering
\includegraphics[width=0.5\textwidth]{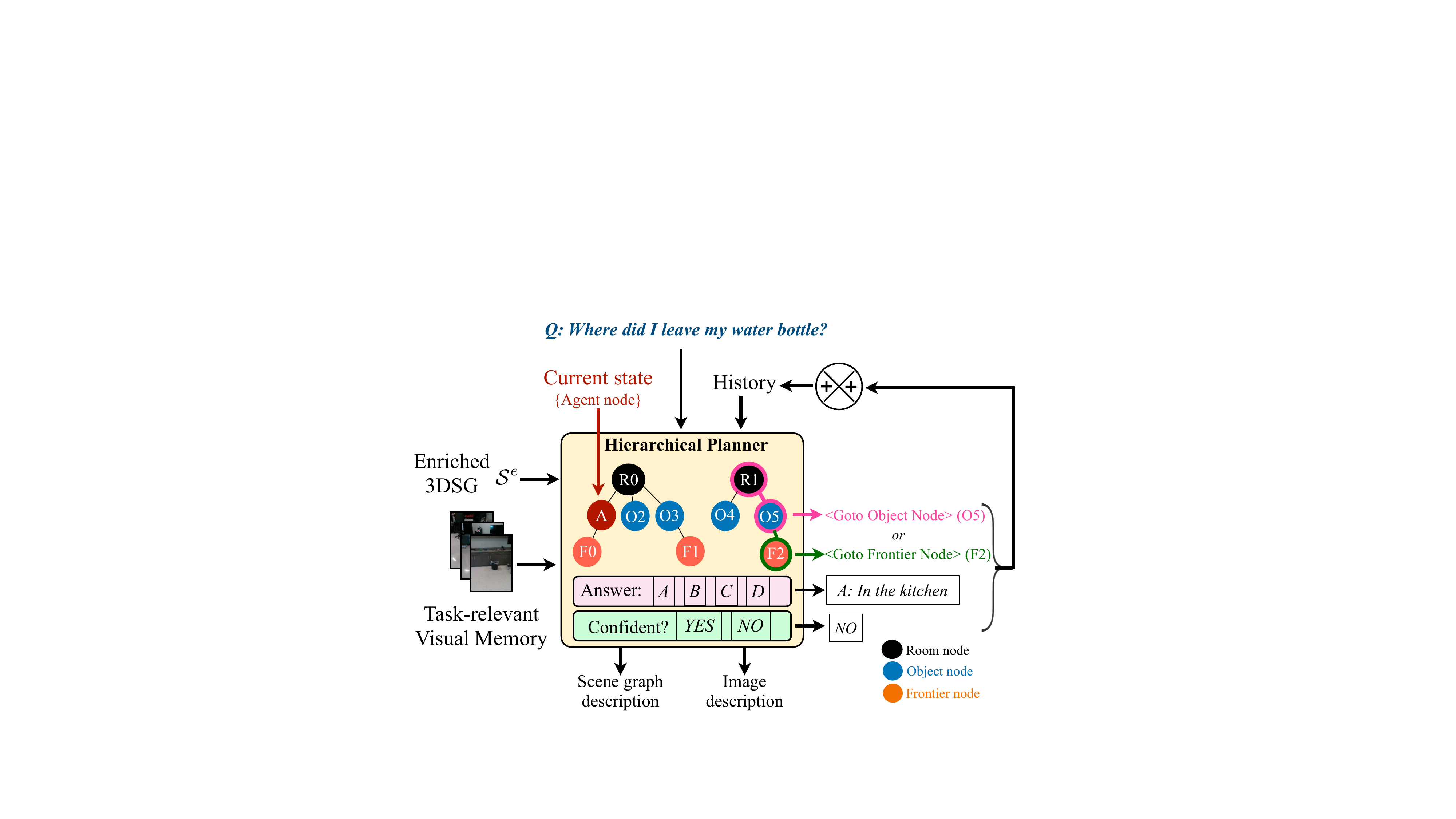}
    \vspace{-2.0em}
    \caption{\footnotesize{\textbf{VLM Planner Architecture.} The Hierarchical Vision-Language planner takes as input the question, enriched scene graph, task-relevant visual memory, current state of the robot (position and room) and a history of past states, actions, answers and confidence values. The planner chooses the next \texttt{<Goto\_Object\_node>} action hierarchically by first selecting the room node and then the object node. The \texttt{<Goto\_Frontier\_node>} action is chosen based on the object nodes connected to the frontier via edges in the scene graph. The planner is asked to output a brief reasoning behind choosing each action, an answer, confidence in its answer, reasoning behind the answer and confidence, the next action, a brief description of the scene graph, and the visual memory.}}
    \label{fig:vlm}
\end{wrapfigure}
\noindent\textbf{Inputs:} At every planning step $t$, the VLM planner takes as input a multiple-choice question $q$, the set of multiple-choice answers $\mathcal{A}$, the enriched scene graph $\mathcal{S}^e_{t}$, and the task-relevant visual memory $\{\mathcal{I}_k\}^K_{k=1}$; $K$ is the number of images in memory. Additionally, we provide the planner with a structured history $\mathcal{H}_{t}$ and the agent's current state $X_t$; $X_t$ is defined in the following format: {\small\texttt{"The agent is currently at node <agent\_node\_id> at position  <agent\_position> in room <room\_id> <room\_name>"}}, where information in `{\small\texttt{<$\cdot$>}}' is populated from $\mathcal{S}^e_{t}$ (see Fig.~\ref{fig:vlm}).\\
\textbf{Outputs:} Given the above inputs, the planner must output an answer $a_{t} \in \mathcal{A}$ for $q$, a boolean value $c_{t} \in \{\text{True}, \text{False} \}$ representing whether it is confident in answering the question, its current numeric confidence level $p_{t}^c \in [0,1]$, and the next action $u_{t}$ that the agent should take. We require the planner to also output the reasoning behind the choice of action and its confidence in the rationale. Finally, the planner is required to plan the next few steps, selecting from two high-level action types: {\small\verb|<Goto_Object_node>(object_id)|} and {\small\verb|<Goto_Frontier_node>(frontier_id)|}, where {\small\verb|object_id|} and {\small\verb|frontier_id|} are selected from $\mathcal{S}^e_{t}$. Selecting an object node enables further visual examination of relevant visited areas. Selecting a frontier node enables visitation of unexplored areas. 
Finally, the planner is required to output a brief description of the scene graph as well as a brief description of the input images. We update the history $\mathcal{H}_{t}$ such that $\mathcal{H}_{t+1}=X_t+a_t+c_t+p_t^c+u_t+\mathcal{H}_t
$.\\
\noindent \textbf{Hierarchical planner and frontier exploration:} For {\small\texttt{<Goto\_Object\_node>(object\_id)}} action types, we enforce a hierarchical planning behavior by requiring the planner to first reason about which room to go to by selecting a room node, then a region node (within the selected room), and finally the object node to go to. This planning behavior reflects the hierarchical structure of the 3DSG and forces the planner to reason about the hierarchical semantics of the scene to explore and answer the questions. For {\small\texttt{<Goto\_Frontier\_node>(frontier\_id)}} action types, we require the planner to provide rationale for its choice of frontier node by referring to the object nodes connected to the selected frontier by edges in the scene graph, enforcing semantic reasoning in the frontier-selection process so that chosen frontiers are task-relevant and for information-gathering. \\
\noindent \textbf{Termination condition:} A planning episode is terminated when the planner outputs $c_{t}=\text{True}$ or $p_{t}^c > 0.9$, i.e., when it is confident in answering the question. The episode is also terminated if $t > T_{max}$, when the maximum allowed planning steps have been reached.\\
\noindent \textbf{Prompt description:} We provide the planner with a system prompt detailing how to understand the scene graph structure, and explain the criteria behind choosing the actions---hierarchically for object nodes and task-relevant or information-gathering for frontier nodes. We explain that the 3DSG can be imperfect/incomplete and that the planner should always seek visual confirmation before answering the question with confidence while employing the scene graph as a semantic map for examining and exploring the scene. Finally, we prompt the VLM to provide a brief description of the scene graph and the input images, focusing on elements in the scene that are relevant to the current task. The complete prompt is available in Appendix \ref{sec:prompting}.

\vspace{-6pt}
\section{Experimental Setup}
\label{sec:exp_setup}

We identify the key research questions that this work aims to evaluate:
\noindent\textbf{Q1)} Do hierarchical 3D scene graphs provide an effective \textbf{metric-semantic memory} for solving \textbf{embodied question answering} tasks? 
\noindent\textbf{Q2)} How does the \textbf{real-time} nature of \ourmethod compare to offline approaches that provide the planner with full-state scene graphs? Specifically, we aim to evaluate if \ourmethod can utilize incrementally constructed state information to solve EQA tasks and terminate based on \textbf{confidence}, without needing to acquire full state information. 
\noindent\textbf{Q3)} Can \ourmethod effectively combine and reason about the \textbf{high-level, semantically-sparse and task-agnostic} information offered by \textbf{scene graphs} and the \textbf{local, semantically-rich and task-relevant} information from \textbf{visual memory} to actively take information gathering actions until it can confidently answer an embodied question?

\subsection{Baselines and Ablations}
\vspace{-4pt}
\label{subsec:baselines}
To evaluate our method and answer the above research questions, we compare against several baselines and focus on methods that employ VLM-based planners for solving EQA or object goal navigation tasks. We compare against a strong baseline, \textbf{Explore-EQA} \cite{ren2024exploreconfident}, which calibrates Prismatic-VLM \cite{karamcheti2024prismatic} to answer embodied questions confidently while maintaining a 2D semantic memory and using prompted images to guide exploration. Note that ExploreEQA is always executed for a pre-specified maximum number of steps, with the highest confidence step chosen to answer the question, while \ourmethod terminates based on a confidence threshold. We implement additional variants of ExploreEQA with newer foundation models---e.g., GPT4o, Llama 4 Maverick, and Gemini 2.5 Pro---for fair comparison with respective VLM variants of \ourmethod.

We also compare \ourmethod against a modified version of \textbf{SayPlan} \cite{rana2023sayplan} which we call \textbf{SayPlanVision}. Similar to SayPlan, SayPlanVision first constructs a scene graph of the whole scene offline and then uses this scene graph for planning. For fair comparison, we further augment SayPlanVision with some abilities of \ourmethod in order to evaluate the effectiveness of our real-time approach and to answer \textbf{Q2}; we provide it with a task-relevant visual memory and confidence-based termination. 

We further evaluate our method for \textbf{Q3}, using two ablations: \textbf{\ourmethod-SG}, where the planner only has access to the real-time 3DSG and does not have access to images; and \textbf{\ourmethod-Vis}, where the planner only takes the visual memory as input and exploration is done via random frontier-based exploration. These ablations will help us evaluate the benefits of multimodality in \ourmethod. 

\noindent\textbf{Experimental Settings:} Since we focus on multi-room environments, we evaluate \ourmethod and the baselines mentioned in \Cref{subsec:baselines} in simulation in Habitat-Sim \cite{szot2021habitat} on scenes from HM3D-Semantics \cite{yadav2022habitat} on the HM-EQA and OpenEQA datasets and in the real-world in two unique home environments. GraphEQA supports open-vocabulary answers and evaluating them using an LLM, as in OpenEQA. However, to ensure consistent evaluation across both benchmarks, we augment the OpenEQA dataset with multiple choice answers using an LLM (see \ref{sec:hmeqa}). We perform numerous experiments with different foundation models as the VLM planner, including GPT4o, Gemini 2.5 Pro, and Llama 4 Maverick. For the real-world setup, we deploy and evaluate our approach on the \textit{Hello Robot Stretch RE-2} mobile manipulation platform with the Stretch AI codebase \cite{stretchai}. All experiments are conducted on a desktop machine with two (2) NVIDIA TITAN RTX GPUs, 64GB of RAM, and an Intel i9-10900K CPU. 

\noindent\textbf{Resource Allocation:} The average token count is broken down as 475 per image (3 images), 1133 for the prompt, and on average 5,425 scene graph tokens, for a total 7983 tokens per VLM step.

\noindent\textbf{Metrics:} We use the following three metrics to compare against the baselines and ablations in ~\Cref{subsec:baselines}: 1) \textit{Success Rate} (\%): an episode is considered a success if the agent answers the embodied question correctly with high confidence; 
2) \textit{Average \# Planning Steps}: For successful episodes, we calculate the average number of VLM planning steps. Note that while Explore-EQA runs for a preset maximum number of steps and post-calculates the maximum confidence step, we report the number of steps taken \textit{until} the max confidence step; 3) \textit{Average Trajectory Length} (meters): for successful episodes, we calculate the average length of the path traveled by the robot. For SayPlanVision, this includes the path traveled to generate the full scene graph. 
\vspace{-4pt}
\subsection{Experimental Results}
\label{subsec:experimental_domains}
\vspace{-8pt}
\label{sec:results}
\begin{table}[h]
\centering
\caption{\footnotesize Comparison to simulation baselines for HM-EQA and OpenEQA datasets: Success rate (\%), average \# of planning steps over successful trials, and $L_\tau$ average trajectory length over successful trials. Methods with a $^\dagger$ indicate our implementations of that particular baseline.}
\renewcommand{\arraystretch}{1.2}
\resizebox{0.95\linewidth}{!}{%
\begin{tabular}{l|ccc|ccc}
\toprule
\multirow{2}{*}{\textbf{Method}} & \multicolumn{3}{c|}{\textbf{HM-EQA}} & \multicolumn{3}{c}{\textbf{OpenEQA}} \\
 & Success Rate (\%) $\uparrow$ & \#Planning Steps $\downarrow$ & $L_\tau$ (m) $\downarrow$ & Success Rate (\%) $\uparrow$ & \#Planning Steps $\downarrow$ & $L_\tau$ (m) $\downarrow$\\
\hline
Explore-EQA \cite{ren2024exploreconfident} & 51.7 & 18.7 & 38.1 & 55.3 & 20.8 & 39.7   \\
Explore-EQA-GPT4o$^\dagger$                & 46.4 & 3.4  & 6.3  & 46.4 & 4.88 & 8.30     \\
Explore-EQA-Llama4-Mav $^\dagger$          & 43.8 & 5.58 & 10.4 & 48.1 & 4.25 & 7.60    \\
Explore-EQA-Gemini-2.5Pro $^\dagger$       & 54.3 & 6.40 & 12.3 & 53.0 & 6.45 & 10.9  \\
SayPlanVision$^\dagger$                    & 54.8 & 2.6  & 5.3  & -  & -  & -          \\
\hline
\ourmethod-GPT4o                           & 63.5 & 5.1 & 12.6 & \textbf{69.1} & 3.97 & 8.29 \\
\ourmethod-Llama4-Mav                      & 57.7 & \textbf{2.36} & \textbf{3.59} & 53.3 & 2.37 & 3.45 \\
\ourmethod-Gemini-2.5Pro                   & \textbf{67.0} & 2.94 & 7.41 & 62.0 & \textbf{2.16} & \textbf{4.03} \\
\bottomrule
\end{tabular}
}
\label{tab:baselines}
\vspace{-5pt}
\end{table}

\para{Comparison to Baselines.} \Cref{tab:baselines} shows simulation results comparing \textbf{\ourmethod} to the baselines discussed in \Cref{subsec:baselines} on the HM-EQA and OpenEQA datasets. Overall, \ourmethod outperforms all other baselines. Compared to \textbf{Explore-EQA}, our method completes tasks in significantly fewer planning steps and with lower trajectory length, indicating more efficient navigation. We also observe that the GPT and Llama variants of Explore-EQA have lower success rates than \textbf{Explore-EQA}, with qualitative results indicating overconfidence in VLMs' predictions (see Appendix \ref{sec:error}), leading to terminating episodes before exploring sufficiently. We note that Explore-EQA's Gemini variant performs better than \textbf{Explore-EQA}, likely due to the inherent spatial reasoning capabilities of Gemini 2.5 Pro. \ourmethod outperforms \textbf{SayPlanVision} even though SayPlanVision has access to the complete scene graph.
For additional OpenEQA results please refer to Appendix \ref{app:aeqa}.
We discuss these results in more detail below.

\para{Baseline And Ablation Study.} Regarding \textbf{Q1}, we observe from \Cref{tab:baselines} that \textbf{\ourmethod} has higher success rate, compared to all \textbf{Explore-EQA} variants, across both HM-EQA and OpenEQA datasets, without the need to build an explicit 2D semantic task-specific memory. This demonstrates the capability of 3DSGs to provide an effective metric-semantic memory for EQA tasks. We also observe that \ourmethod requires a significantly lower number of planning steps as compared to \textbf{Explore-EQA}. This is because, unlike \textbf{Explore-EQA}, \ourmethod does not entirely rely on images as input to the VLM planner for building the semantic memory as well as planning, as this would constrain the planner to choose from only regions that are visible in the current image. \textbf{\ourmethod}, on the other hand, can use the hierarchical structure of the scene graph as well as semantically-enriched frontier nodes to plan across the entire explored space. 


Additional error analysis in Appendix \ref{sec:error} reveals that the GPT variant of Explore-EQA has a significantly high percentage of \textit{false positives}, i.e., questions that are answered successfully using commonsense reasoning/guessing, but without grounding the answer in the current scene.
This provides additional evidence of the effectiveness of 3DSGs in enabling \textit{semantic exploration} by grounding the planner in the current environment. Qualitatively, we observe that actions chosen by the planner, {\footnotesize\texttt{<Goto\_Object\_node>(object\_id)}} and {\footnotesize\texttt{<Goto\_Frontier\_node>(frontier\_id)}},  are task-specific and explainable. For more qualitative results please refer to Appendix \ref{app:qual}. 

For \textbf{Q2}, we observe from~\Cref{tab:baselines} that \ourmethod performs better than \textbf{SayPlanVision} which has access to the complete scene graph. This is a surprising result since it is expected that, given full scene graph information, SayPlanVision would outperform \ourmethod across all metrics. However, from a qualitative analysis of the results for \textbf{SayPlanVision}, we observe that given access to the complete scene graph, the context is too large, providing for a much harder inference problem for the VLM. As a result, the agent is overconfident about its choice of object node actions, leading to shorter trajectory lengths in successful cases, but also to increased failure cases. This points to an interesting advantage of our real-time approach---that \textit{incrementally} building memory by exploring task-relevant regions and maintaining a more concise representation benefits EQA tasks.

For \textbf{Q3}, the ablation results in~\Cref{tab:ablations} for \textbf{\ourmethod-SG} show lower success rate, higher average planning steps, and higher average trajectory length, compared to \ourmethod, as it only uses the 3DSG as textual input to the VLM planner, demonstrating that a semantic scene graph constructed using closed-set semantics and without any task-specific semantic enrichment will provide an incomplete and insufficient environment understanding; \ourmethod's task-relevant visual memory and task-specific enrichment are crucial for solving EQA tasks. Furthermore, we note that the performance of the vision-only ablation \textbf{\ourmethod-Vis} \textit{also} suffers: this is because the agent takes random exploratory actions in the environment, with no semantic understanding of the scene structure to guide exploration. However, qualitatively we observe that  without access to a scene graph to ground the agent in the current environment, \ourmethod-Vis exhibits overconfidence (taking very few planning steps) and tries to answer the question solely based on the current image.
\begin{wraptable}{r}{0.5\textwidth}
  \centering
  \vspace{-10pt}
  \caption{\footnotesize Ablations (Simulation): Success rate (\%), number of planning steps and $L_\tau$ the trajectory length.}
  \label{tab:ablations}
  \setlength{\tabcolsep}{7pt} 
  \resizebox{0.5\columnwidth}{!}{
  \begin{tabular}{@{}lccc@{}}
    \toprule
    Ablation & Succ. Rate (\%)  & \#Planning steps & $L_\tau$ (m) \\
    \midrule
    \ourmethod-SG & 13.6 & 8.8 & 33.0 \\
    \ourmethod-Vis & 45.7 & 1.0 & 3.9\\
    \ourmethod & \textbf{63.5} & 5.1 & 12.6 \\
    \bottomrule
  \end{tabular}
  }
\end{wraptable}
\textbf{\ourmethod} outperforms all ablations, providing clear evidence on the utility of a multimodal approach that combines global spatial and semantic information from 3D scene graphs with local but rich semantic information from images, for solving challenging EQA tasks. We also observe from ~\Cref{tab:baselines} that \ourmethod exhibits lower average planning steps and lower average trajectory length, while retaining a higher success rate, highlighting the capability of \ourmethod to reason about multimodal data in the form of semantically sparse and task-agnostic scene graphs and more semantically rich information in the form of task-relevant visual memory. Additional ablations are available in Appendix \ref{sec:app_add_ablations}. 

\begin{figure*}[t]
    \centering
\includegraphics[width=0.96\textwidth]{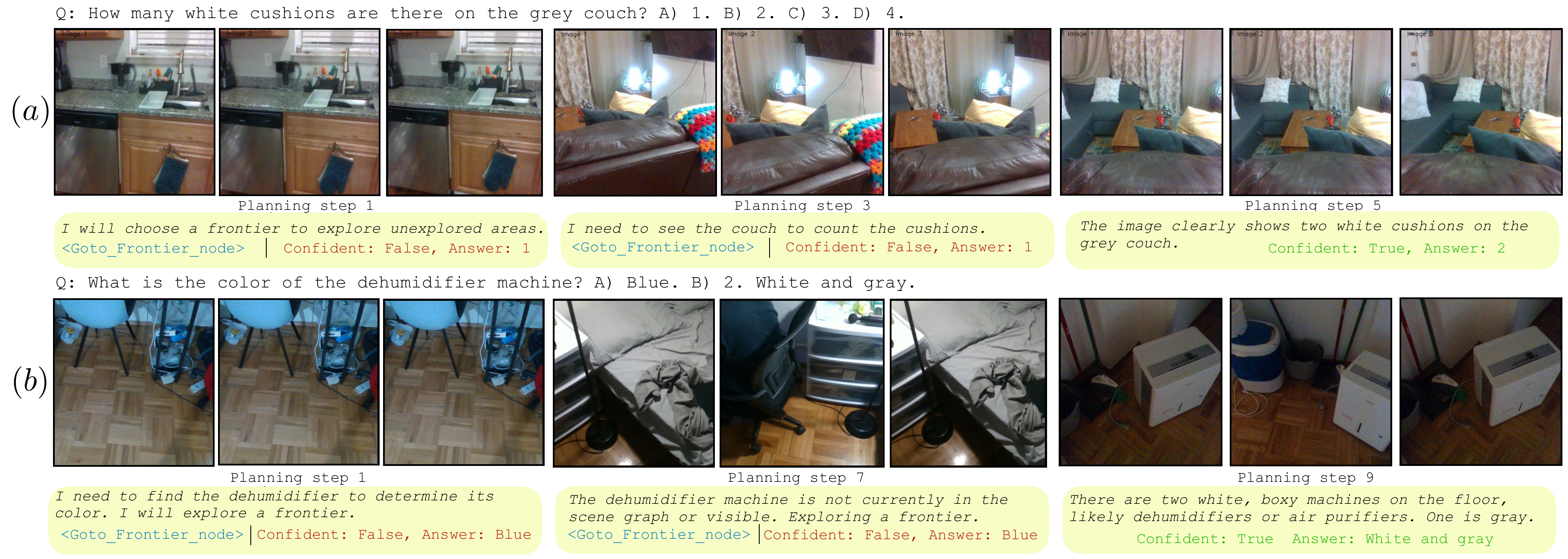}
    \vspace{-7pt}
    \caption{\footnotesize{\textbf{Images from real-world experiments}, deploying \ourmethod on the Hello Robot Stretch RE2 platform in two unique home environments (a, b). Each set of images is from the head camera on the Stretch robot, representing the top-K task-relevant images at each planning step as it constructs the scene graph and attempts to answer the question with high confidence. Provided under the images are planning step, answers, confidence, and explanations output from the VLM planner.}}
    \vspace{-15pt}
    \label{fig:real_env_01}
\end{figure*}

\para{Real-world Experiments.}
We deploy \ourmethod on the \textbf{Hello Robot Stretch RE2} platform across two home environments. We conduct ten trials in Home (a) and five trials in Home (b).~\Cref{fig:real_env_01} illustrates representative runs from each setting. For each environment, we design a custom set of Embodied Question Answering (EQA) tasks (see Appendix~\ref{app:questions} and \href{https://saumyasaxena.github.io/grapheqa/}{website}), aligned with the task categories described in Appendix~\ref{sec:hmeqa}. To construct the underlying 3D metric-semantic scene graph, \ourmethod employs Detic~\cite{zhou2022detic} for semantic segmentation and integrates RGB-D images along with camera intrinsics and extrinsics as input to the Hydra scene graph generator~\cite{hughes2022hydra}.
\vspace{-10pt}

\section{Conclusion}
\vspace{-2pt}
We present \ourmethod, an approach for solving embodied question answering tasks in unseen environments by grounding a vision-language based planner in the context of the current environment by providing as input textual representations of real-time 3D metric-semantic scene graphs and a task-relevant visual memory. We show that \ourmethod achieves improved performance over existing approaches on EQA tasks in both the HM-EQA and OpenEQA benchmarks, and validate performance across both \textit{closed} and \textit{open} multimodal foundation models. Furthermore, we validate its practical applicability through real-world deployments in two indoor environments.
\newpage
\section{Limitations}
A limitation of this approach is reliance on off-the-shelf segmentation and detection models for creating semantic maps required for 3DSG construction. The performance of our approach, hence, is directly impacted by the performance of the detection method used and the semantic categories in the scene graph are limited to the categories detected by the segmentation model. An exciting direction for future work includes enriching the scene graph online with task-relevant node and edge features using open-set models. Another limitation of our approach is that VLM-based planners can be overconfident or underconfident when answering embodied questions. Grounding calibrated models in real-time multi-modal memory is another important direction for future work. Additionally, \ourmethod is currently limited to single-floor scenes, and is unable to traverse multiple floors or buildings. Accommodating multi-floor scenes is within the scope of future work.
\bibliography{example}  
\clearpage
\setcounter{page}{1}
\appendix

\renewcommand{\thesection}{\Alph{section}} 
\renewcommand{\thesubsection}{\thesection.\arabic{subsection}} 

\label{sec:appendix}

\section{Experiment setup}
\subsection{Habitat environment setup}
\label{sec:habitat}

The Habitat-Sim setup for our experiments is identical to the setup used in \cite{ren2024exploreconfident}. The camera sensor settings are as follows: image width = 640, image height = 480, camera height = 1.5m, camera tilt = $-30$ degrees, field of view = 120 degrees.
For generating trajectories for the {\small\verb|<Goto_Object_node>(object_id)|} and {\small\verb|<Goto_Frontier_node>(frontier_id)|} actions, we find the shortest path between the current agent position and the desired object/frontier node location, on the obstacle-free voxel space of of the 2D occupancy map. We orient the agent such that camera always points towards the desired node location all along the trajectory. In case of the {\small\verb|<Goto_Object_node>(object_id)|} action, this maximizes the number of views that capture the target object. In case of the {\small\verb|<Goto_Frontier_node>(frontier_id)|} action, this makes the agent look outwards into the unexplored areas. 

\subsection{HM-EQA and OpenEQA Benchmarks}
\label{sec:hmeqa}
\para{HM-EQA:} The Habitat-Matterport Embodied Question Answering (HM-EQA) dataset introduced by Ren et al. \cite{ren2024exploreconfident} is based in the Habitat-Matterport 3D Research Dataset (HM3D) of photo-realistic, diverse indoor 3D scans \cite{ramakrishnan2021hm3d}. The dataset is composed of 500 multiple choice questions from 267 different scenes which fall in the following categories: identification, counting, existence, state, and location. A sample from the \textit{identification} category is as follows: \texttt{Q: Which rug did I put next to the kitchen sink? A. There is no rug. B. White one. C. Gray one. D. Green one.}
\\
\para{OpenEQA:} The OpenEQA benchmark \cite{OpenEQA2023} is composed of data for two different settings, \textit{EM-EQA} (episodic memory) and \textit{A-EQA} (active exploration), with questions spanning seven categories, spatial understanding, object state recognition, functional reasoning, attribute recognition, world knowledge, object localization, and object recognition, in two different environments, HM3D and ScanNet. 
The active exploration dataset (A-EQA) consists of 557 questions and corresponding open vocabulary answers, for example, \texttt{Q: Do I have Cayenne pepper left at home? A: I found a bottle of Cayenne pepper in the pantry}. In this setting, the agent is supposed to explore the environment autonomously to answer the question. The episodic memory dataset (EM-EQA) consists of a sequence of historical sensory (RGB) observations in addition to each question-answer pair. In this setting, the agent is supposed to analyze the episodic memory to answer the question. To evaluate for success, an LLM is used to score the agent's answer based on similarity to the ground truth answer in the dataset. Evidently, the A-EQA dataset is relevant to our \ourmethod setting, where the agent needs to explore an \textit{unseen} environment to solve an EQA task. We consider A-EQA questions only in the HM3D environment, resulting in a total of 113 questions. We exclude ScanNet questions, as ScanNet is predominantly composed of single-room environments and therefore does not present a substantial exploration challenge for \ourmethod. Furthermore, although GraphEQA supports querying natural language answers from the VLM planner (instead of multiple choice answers) and evaluating them using an LLM, as in OpenEQA; to ensure consistent evaluation to HM-EQA, we augmented the A-EQA dataset with multiple choice answers using an LLM. The correct answer is provided as one of the four choices.

\subsection{3D Scene Graph Construction}
\label{sec:hydra_sgs}
We use Hydra~\cite{hughes2022hydra} to construct a layered metric-semantic scene graph. Hydra 3DSGs are comprised of the following layers: \textit{Layer 1 (bottom)}: a metric-semantic 3D mesh, \textit{Layer 2}: objects with corresponding semantic labels and the agent, \textit{Layer 3}: regions or places, \textit{Layer 4}: rooms, and \textit{Layer 5 (top)}: building. Intra-layer edges between nodes denote `traversability', while inter-layer edges denote `belonging'. For example, an edge between regions in Layer 3 denotes traversability between these regions and an edge between an object and a room denotes that the object is located in that room. 3DSGs are constructed using RGB and depth images, semantic segmentation masks, camera extrinsics, and intrinsics.
The HM3D-Semantics dataset provides ground-truth semantic segmentation masks for simulation experiments.
In real-world experiments, we use Detic \cite{zhou2022detic} to obtain segmentation masks.

\section{Prompting}
\label{sec:prompting}
\subsection{GPT Prompt}
The full prompt provided to GPT4o for \ourmethod is given below. In it we provide the scene graph description, description of the agent's current state, agent prompt, and just generally more descriptive text for more context.

\subsection{Exploiting Hierarchical Nature of 3DSGs for Planning}
The portion of the prompt used to describe the scene graph in \ourmethod clarifies to the VLM how layers and nodes are organized in a 3DSG. We take advantage of this structure by requiring {\small\verb|<Goto_object_node_step>|} to be organized such that the VLM first chooses a room (level 4) to navigate to, then choosing an object (level 2) in that room. This inherent structure and explanation of it in the prompt guides the VLM to choose actions that investigate objects that are semantically relevant to the question.

\newpage
\section*{Full VLM Prompt}

\begin{minipage}{1.1\textwidth}
\begin{tcolorbox}[mymonobox]
\textcolor{blue}{\textbf{Agent prompt}}: You are an excellent hierarchical graph planning agent. 
Your goal is to navigate an unseen environment to confidently answer a multiple-choice question about the environment.
As you explore the environment, your sensors are building a scene graph representation (in json format) and you have access to that scene graph.\\ 
\textcolor{blue}{\textbf{Scene graph description}}: A scene graph represents an indoor environment in a hierarchical tree structure consisting of nodes and edges/links. There are six types of nodes: building, rooms, visited areas, frontiers, objects, and agent in the environment.
The tree structure is as follows: At the highest level 5 is a 'building' node.
At level 4 are room nodes. There are links connecting the building node to each room node.
At the lower level 3, are region and frontier nodes. 'region' node represent region of room that is already explored. Frontier nodes represent areas that are at the boundary of visited and unexplored areas. There are links from room nodes to corresponding region and frontier nodes depicted which room they are located in.
At the lowest level 2 are object nodes and agent nodes. There is an edge from region node to each object node depicting which visited area of which room the object is located in.
There are also links between frontier nodes and objects nodes, depicting the objects in the vicinity of a frontier node.
Finally the agent node is where you are located in the environment. There is an edge between a region node and the agent node, depicting which visited area of which room the agent is located in.\\
\textcolor{blue}{\textbf{Current state description}}: CURRENT STATE will give you the exact location of the agent in the scene graph by giving you the agent node id, location, room\_id and room name.\\
\textcolor{blue}{\textbf{General Description}}: Given the current state information, try to answer the question. Explain the reasoning for selecting the answer.
Finally, report whether you are confident in answering the question. 
Explain the reasoning behind the confidence level of your answer. Rate your level of confidence. 
Provide a value between 0 and 1; 0 for not confident at all and 1 for absolutely certain.
Do not use just commonsense knowledge to decide confidence. 
Choose TRUE, if you have explored enough and are certain about answering the question correctly and no further exploration will help you answer the question better. 
Choose 'FALSE', if you are uncertain of the answer and should explore more to ground your answer in the current environment. 
Clarification: This is not your confidence in choosing the next action, but your confidence in answering the question correctly.
If you are unable to answer the question with high confidence, and need more information to answer the question, then you can take two kinds of steps in the environment: Goto\_object\_node\_step or Goto\_frontier\_node\_step 
You also have to choose the next action, one which will enable you to answer the question better. 
Goto\_object\_node\_step: Navigates near a certain object in the scene graph. Choose this action to get a good view of the region around this object, if you think going near this object will help you answer the question better. Important to note, the scene contains incomplete information about the environment (objects maybe missing, relationships might be unclear), so it is useful to go near relevant objects to get a better view to answer the question. 
Use a scene graph as an imperfect guide to lead you to relevant regions to inspect.
Choose the object in a hierarchical manner by first reasoning about which room you should goto to best answer the question, and then choose the specific object.
Goto\_frontier\_node\_step: If you think that using action ``Goto\_object\_node\_step'' is not useful, in other words, if you think that going near any of the object nodes in the current scene graph will not provide you with any useful information to answer the question better, then choose this action.
This action will navigate you to a frontier (unexplored) region of the environment and will provide you information about new objects/rooms not yet in the scene graph. It will expand the scene graph. 
Choose this frontier based on the objects connected this frontier, in other words, Goto the frontier near which you see objects that are useful for answering the question or seem useful as a good exploration direction. Explain reasoning for choosing this frontier, by listing the list of objects (<id> and <name>) connected to this frontier node via a link (refer to scene graph).

While choosing either of the above actions, play close attention to 'HISTORY' especially the previous 'Action's to see if you have taken the same action at previous time steps. 
Avoid taking the same actions you have taken before.
Describe the CURRENT IMAGE. Pay special attention to features that can help answer the question or select future actions.
Describe the SCENE GRAPH. Pay special attention to features that can help answer the question or select future actions.''
\end{tcolorbox}

\begin{center}
The prompt used in the implementation of \ourmethod.
\end{center}
\end{minipage}

\subsection{Structured Output}
We employ the structured output capabilities of OpenAI's Python API to force a desired structure on what is output by GPT4o. Below is the \texttt{create\_planner\_response} function used in the implementation of \ourmethod.

\begin{minipage}{\textwidth}
\begin{lstlisting}[caption=The \texttt{create\_planner\_response} function used to structure output from GPT4o., label={lst:listing-python}, language=Python]
def create_planner_response(frontier_node_list, room_node_list, region_node_list, object_node_list, Answer_options):

    class Goto_frontier_node_step(BaseModel):
        explanation_frontier: str
        frontier_id: frontier_node_list

    class Goto_object_node_step(BaseModel):
        explanation_room: str
        explanation_obj: str
        room_id: room_node_list
        object_id: object_node_list
    
    class Answer(BaseModel):
        explanation_ans: str
        answer: Answer_options
        explanation_conf: str
        confidence_level: float
        is_confident: bool

    class PlannerResponse(BaseModel):
        steps: List[Union[Goto_object_node_step, Goto_frontier_node_step]]
        answer: Answer
        image_description: str
        scene_graph_description: str

    return PlannerResponse
\end{lstlisting}
\end{minipage}

The \texttt{create\_planner\_response} function takes as input enums for frontier nodes, room nodes, region nodes, object nodes, and the answer options for the particular question being answered by the VLM. These enums are used to populate the member variables of the \texttt{Goto\_frontier\_node\_step}, \texttt{Goto\_object\_node\_step}, and \texttt{Answer} classes,  enforcing both type as well as the options available when calling the OpenAI API.

\newlist{questions}{enumerate}{1}
\setlist[questions,1]{
  label=Q\arabic*.,      
  labelsep=0.5em,        
  leftmargin=*,          
  itemsep=0pt,           
  parsep=0pt,            
  topsep=0pt,            
  partopsep=0pt
}

\section{Additional Simulation Experiments}

\subsection{Performance across Task Categories}
\label{sec:app_categories}

\begin{table}[h]
  \centering
  \caption{\footnotesize Success Rate (\%) in simulation for task categories in the HM-EQA dataset, for Identification, Counting, Existence, State, and Location categories. Reported in terms of category successes / total number of category EQA tasks. $^\dagger$ indicates our implementation of that baseline.}
  \label{tab:succcategory}
  \setlength{\tabcolsep}{4pt} 
  \resizebox{0.6\columnwidth}{!}{
  \begin{tabular}{@{}lccccc@{}}
    \toprule
      Method & \footnotesize{Ident.}  & \footnotesize{Counting} & \footnotesize{Existence} & \footnotesize{State} & \footnotesize{Location} \\
    \midrule
    \footnotesize Explore-EQA & 59.2 & 46.2 & 56.5 & 46.5 & 47.7 \\
    \footnotesize Explore-EQA-GPT4o$^{\dagger}$ & 32.5 & 44.2 & 56.4 & 42.3 & 40.8 \\
    \footnotesize SayPlanVision$^{\dagger}$ & 75 & 44.4 & 63.3 & 43.4 & 56\\
    \footnotesize \ourmethod & \textbf{77.8} & \textbf{57.9} & \textbf{69} & \textbf{65.2} & \textbf{64} \\
    \bottomrule
  \end{tabular}
  }
\end{table}

~\Cref{tab:succcategory} shows the performance of baselines and \ourmethod across the different task categories (identification, counting, existence, state, location)  in the HM-EQA dataset. 
\ourmethod outperforms all other methods across all task categories, but is particularly more performant in comparison when considering \textit{Counting} and \textit{State} tasks. It is worth noting that the \textit{Counting} and \textit{State} categories are among the most challenging. 

\subsection{Additional Ablations}
\label{sec:app_add_ablations}
We perform some additional ablations to evaluate the utility of different components of our method: 
\textbf{\ourmethod-NoEnrich}, which does not use frontier   enrichment (\Cref{enrichment}), and \textbf{\ourmethod-CurrView}, which uses only the current view as input to the VLM and does not choose additional task-relevant keyframes (\Cref{keyframe}). All ablations of GraphEQA use GPT-4o.
Here we analyze these two additional ablations, \textbf{\ourmethod-NoEnrich} and \textbf{\ourmethod-CurrView}. We observe that \ourmethod-NoEnrich performs slightly worse than \ourmethod which demonstrates that enriching the scene graph with additional semantic information in the form of edges between frontiers and nearest objects, as discussed in ~\Cref{enrichment}, lends itself to semantically informed exploration. We observe that the performance drop is worse in the case of \ourmethod-CurrView, where we do not use task-relevant visual memory, but only the current view of the agent. This demonstrates that task-relevant visual memory is very useful in long-horizon tasks where the current view of the robot might not be the best view for answering an embodied question.

\begin{table}[h]
  \centering
  \caption{\footnotesize Ablations (Simulation): Success rate (\%), number of planning steps and $L_\tau$ the trajectory length. }
  \setlength{\tabcolsep}{4pt} 
  \resizebox{0.6\columnwidth}{!}{
  \begin{tabular}{@{}lccc@{}}
    \toprule
    Ablation & Succ. Rate (\%)  & \#Planning steps & $L_\tau$ (m) \\
    \midrule
    \ourmethod-NoEnrich & 59.5 & 5.1 & 11.1 \\
    \ourmethod-CurrView & 53.1 & 5.7 & 12.2 \\
    \ourmethod & \textbf{63.5} & 5.1 & 12.6 \\
    \bottomrule
  \end{tabular}
  }
  \label{tab:add_ablations}
\end{table}

\subsection{Additional Quantitative Analysis: Zero-shot reasoning for commonsense questions}
\label{sec:zeroshot}
We perform this evaluation to answer the following question: how many questions in the HM-EQA dataset can be answered \textit{correctly} purely based on commonsense reasoning or guessing, without exploring the environment?
We aim to perform this analysis to roughly estimate the percentage of false positives that can occur in the HM-EQA dataset using different VLMs. 
To answer this question we define the following three additional baselines: \textbf{EQAZeroShotGPT4o}, \textbf{EQAZeroShotPrismatic} and 
\textbf{EQAZeroShotGPT4oQuestionOnly}. \textbf{EQAZeroShotGPT4o} evaluates the zero-shot performance of GPT-4o when answering an EQA question using the initial image and the question. \textbf{EQAZeroShotPrismatic} evaluates the zero-shot performance of the calibrated Prismatic model from Explore-EQA \cite{ren2024exploreconfident} when answering an EQA question using the initial image and the question. \textbf{EQAZeroShotQuestionOnly} evaluates the zero-shot performance of GPT-4o when answering an EQA question using only the question. In all the above baselines, no exploration steps are taken. Prompts for the above baselines are identical to ones used by Explore-EQA \cite{ren2024exploreconfident}. An episode is considered a success if the question is answered correctly and with high confidence ($>0.5$). 

\Cref{tab:additional_ablations} shows the simulation results for the baselines mentioned above compared to the baselines discussed in \Cref{subsec:baselines}. We observe that, given only the question, \textbf{EQAZeroShotQuestionOnly} answers $6.6 \%$ of the questions correctly with high confidence. This can be attributed to the VLM answering questions based on commonsense reasoning or even just random guessing, and getting them correct.
\textbf{EQAZeroShotPrismatic} answers only $1.8\%$ of the questions correctly with a confidence $>0.5$. This can be attributed to the fact that the Prismatic VLM is calibrated to avoid overconfident behavior in the absence of evidence, hence the zero-shot confidence values are low. \textbf{EQAZeroShotGPT4o} answers $17.2 \%$ of the questions correctly with high confidence. To evaluate whether these questions were answered based on actual evidence in the initial image or purely based on commonsense reasoning/guessing, we further qualitatively evaluated the successful cases. Among the $17.2 \%$ that \textbf{EQAZeroShotGPT4o} answers successfully, $8.8 \%$ were answered based on actual evidence in the initial image. These questions could be answered using the initial image. The remaining $8.4 \% $ were answered based on commonsense reasoning/guessing without any evidence from the environment. Thus, $8.4 \% $ is the rough estimate of the false positives that can occur in the HM-EQA dataset using GPT-4o based methods.

\begin{table}
  \centering
  \caption{Additional baselines (Simulation): Success rate (\%)}
  \setlength{\tabcolsep}{6pt} 
  \resizebox{0.6\columnwidth}{!}{
  \begin{tabular}{@{}lc@{}}
    \toprule
    Method & Succ. Rate (\%) \\
    \midrule
    Explore-EQA \cite{ren2024exploreconfident} & 51.7  \\
    Explore-EQA-GPT4o  & 46.4 \\
    SayPlanVision & 54.8 \\
    \ourmethod & \textbf{63.5} \\
    EQAZeroShotGPT4o & 17.2 \\
    EQAZeroShotPrismatic & 1.8  \\
    EQAZeroShotGPT4oQuestionOnly & 6.6 \\
    \bottomrule
  \end{tabular}
  }
  \label{tab:additional_ablations}
\end{table}

\subsection{Error Analysis of Competing Baselines}
\label{sec:error}
\begin{figure*}[h]
    \centering
\includegraphics[width=\textwidth]{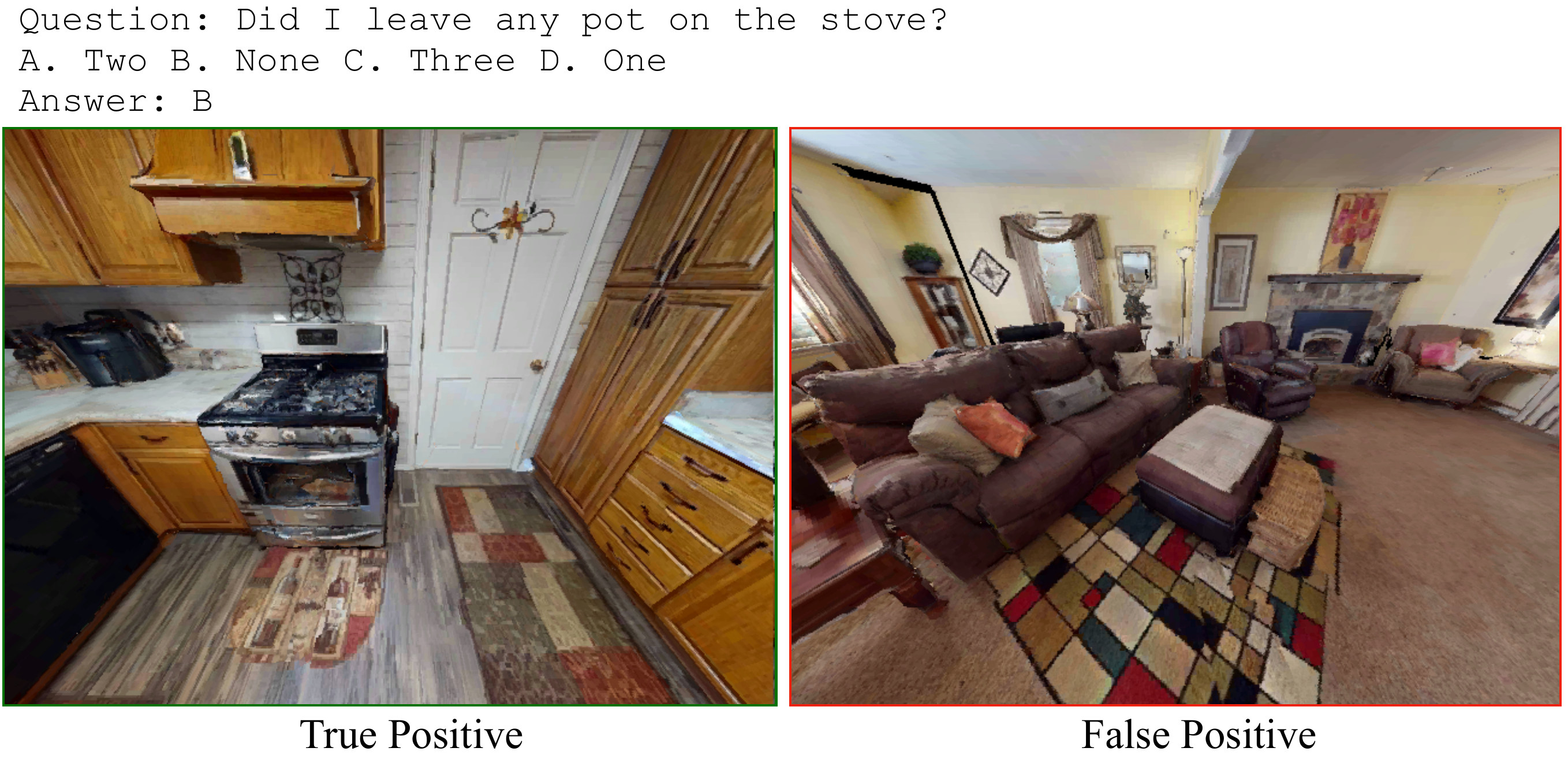}
    \caption{\footnotesize{An example of a false positive case. The image on the left is the image that can be used to answer the question correctly. The image on the right is the image used by an agent to 'guess' the answer correctly with high confidence without grounding the answer in the current environment.}}
    \label{fig:false_pos}
\end{figure*}

\begin{figure*}[h]
\centering
\includegraphics[width=\textwidth]{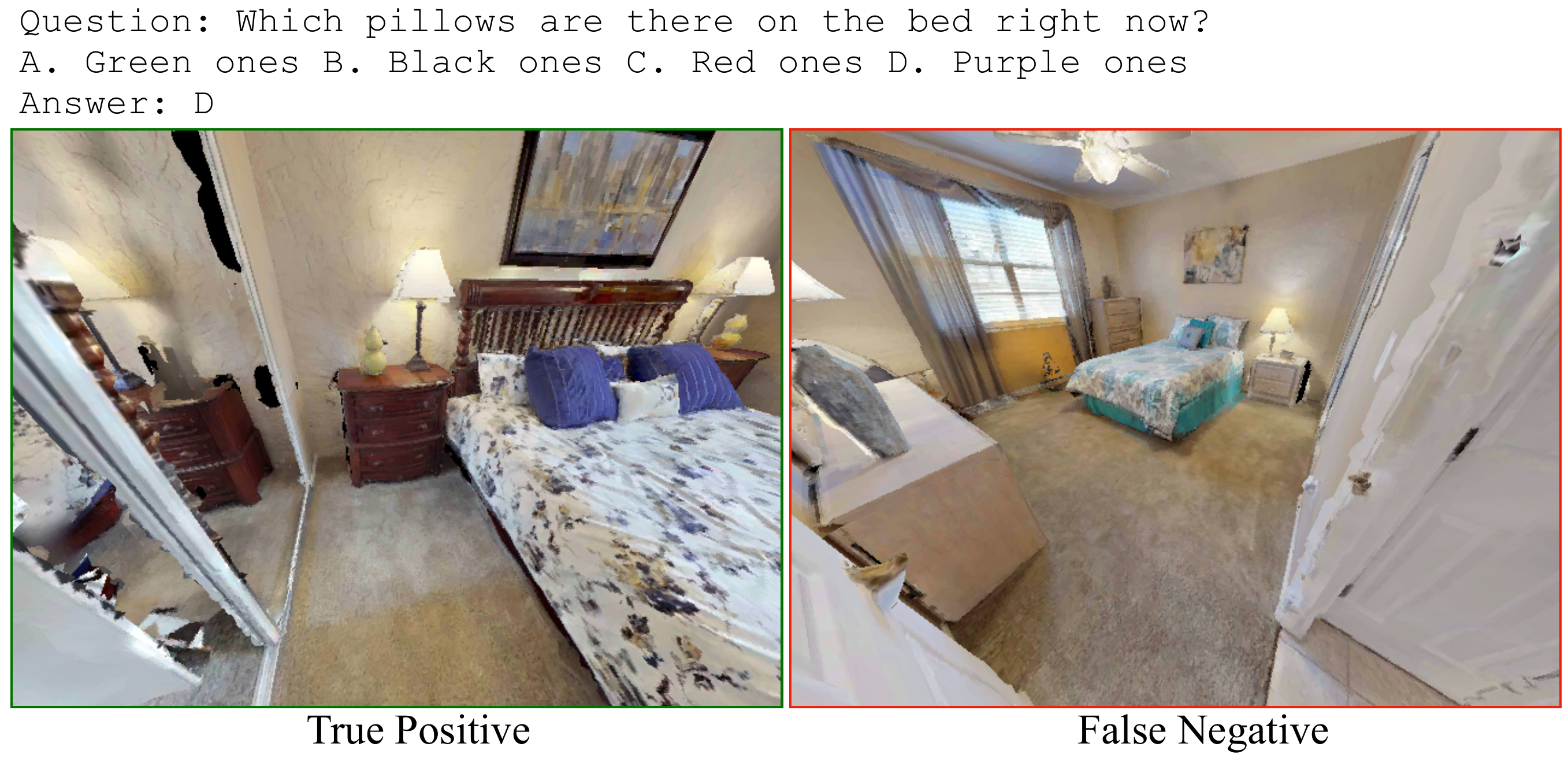}    \caption{\footnotesize{An example of a false negative case. The question inquires about the color of the pillow on the bed. The question is ambiguous. On the left is the image that corresponds to the answer in the dataset i.e. purple pillows. On the right is an image that the agent encounters during exploration and answers that the pillows are 'green' with high confidence. Given the image, the answer is correct but is deemed incorrect in the dataset.}}
    \label{fig:false_neg}
\end{figure*}

Given the nature of the EQA tasks, it is possible that some of the questions are answered successfully using only commonsense reasoning/guessing, without grounding the answer in the current environment. We consider these cases as \textit{\textbf{false positives}}. An example of a false positive is shown in \Cref{sec:appendix} \Cref{fig:false_pos}.  Furthermore, we also notice \textit{\textbf{false negatives}}, where the answer was marked incorrect given the answer in the data set, although given the current image and scene graph, the answer seemed appropriate. Such cases exist due to ambiguities in the dataset. An example of a false negative is shown in \Cref{fig:false_neg}. To get an estimate of the number of false positives and false negatives in our baselines, we consider a set of 114 questions from the HM-EQA dataset and manually label the results across the four categories: True Positives, True Negatives, False Positives, and False Negatives. The results are shown in \Cref{tab:confusion_matrix} where each number is a percentage of the total number of questions considered (114). We limit this error analysis to only the GPT variant, as it is performed manually by a human.

\begin{table}[h] 
  \centering
  \caption{\footnotesize Error analysis (Simulation): Percentage $\%$}
  \label{tab:confusion_matrix}
  \setlength{\tabcolsep}{8pt} 
  \resizebox{0.6\columnwidth}{!}{
  \begin{tabular}{@{}lccc@{}}
    \toprule
     & \ourmethod & Explore-EQA & Explore-EQA-GPT4o \\
    \midrule
    True positive & \textbf{58.18} & 31.58 & 22.81 \\
    True negative  & \textbf{31.82} & 44.74 & 46.49 \\
    False positive & \textbf{6.36} & 16.67 & 24.56 \\
    False negative & \textbf{3.64} & 7.02 & 6.14\\
    \bottomrule
  \end{tabular}
  }
\end{table}

From \Cref{tab:confusion_matrix}, we observe that \ourmethod has the least number of false positives and false negatives, yielding more reliable success rates. We note that \textbf{Explore-EQA-GPT4o} has a considerable percentage of false positives, i.e., questions are answered correctly based on guessing without grounding the answer in the current environment. This sheds light on why \textbf{Explore-EQA-GPT4o} has comparable success rates to \textbf{Explore-EQA-GPT4o}, even with considerably fewer planning steps (\Cref{tab:baselines}).
This provides further evidence that \ourmethod effectively grounds GPT-4o in the current environment, is not overconfident based on commonsense reasoning and explores the environment until it can answer the question based on evidence. See additional results in \Cref{sec:app_categories}, categorized by question type.

\subsection{Comparison to OpenEQA baselines on the A-EQA dataset}
\label{app:aeqa}
As discussed in \ref{sec:hmeqa}, to ensure consistent evaluation across both simulation benchmarks, we augmented the A-EQA dataset with multiple-choice answers using an LLM.
In the original A-EQA setting (with open-vocabulary responses and LLM-based evaluation), GraphEQA-GPT4o achieves a success rate of 53.6\%, which, as expected, is lower than the multiple choice setting given the increased difficulty of the open-vocabulary task.
\\
\para{Comparison to Multi-Frame VLM} - 
We compare our result above, to OpenEQA's best performing method  in the A-EQA setting -- \textbf{Multi-Frame VLM}.
Since code for A-EQA experiments is not released (only for EM-EQA), and some implementation details such as the random exploration strategy and termination condition are not specified, we refer to the results reported in the paper, where Multi-Frame VLM achieves a success rate of 41.8\%.
We attribute the performance gap between \ourmethod and \textbf{Multi-Frame VLM} to: (1) Multi-Frame VLM uses random exploration, whereas GraphEQA explores via semantic guidance; (2) Multi-Frame VLM's visual memory consists of 50 \textit{uniformly sampled} frames, which risks missing critical task-relevant images in large HM3D scenes (hundreds of images); and (3) as evidenced by comparisons to SayPlanVision, 
compact task-relevant memory, as used in \ourmethod, outperforms large-context inputs (large number of images in Multi-Frame VLM), since overly large contexts can degrade VLM performance.

\subsection{Qualitative Analysis}
\label{app:qual}
\subsubsection{Exploration Efficiency and Trajectory Generation}
\begin{figure*}[h]
    \centering
\includegraphics[width=0.9\textwidth]{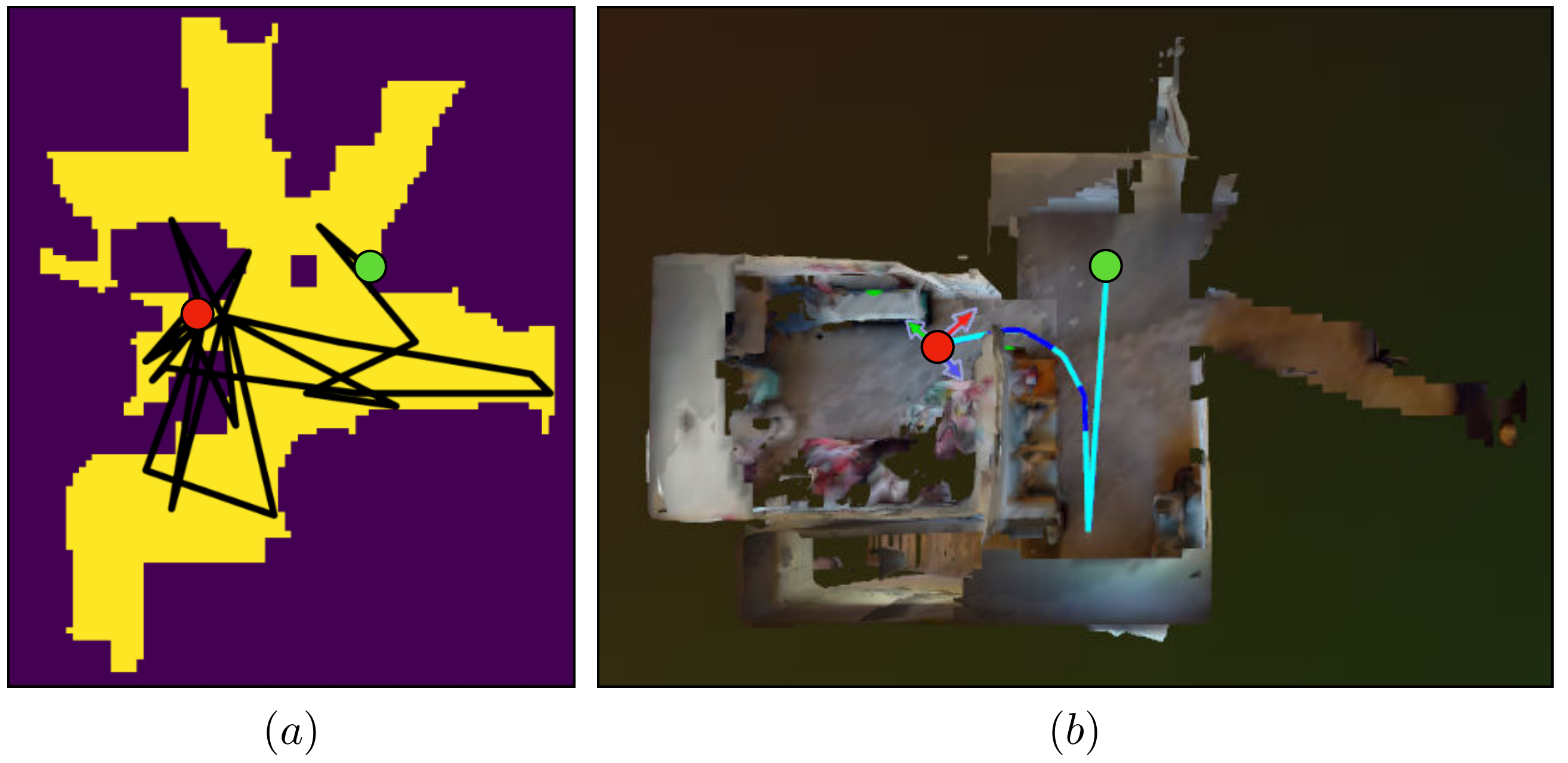}
    \caption{(a) A top down view of the map constructed by Explore-EQA illustrating explored areas for a scene in the HM3D dataset. The agent's initial position is depicted in green and its final position in red. The trajectory of the agent is shown in black. (b) A top down view of scene reconstruction output by  as GraphEQA explores the same HM3D scene used in the Explore-EQA map to the left. The agent's initial position is again in green and its final position in red. The agent's trajectory is shown in blue in this image. In this figure, we want to bring attention to the contrast between a typical trajectory taken by an agent employing GraphEQA vs. a typical trajectory taken by an agent employing Explore-EQA.}
    \label{fig:exploration_eff}
\end{figure*}

We illustrate the differences in exploration between GraphEQA and Explore-EQA through the following example. ~\Cref{fig:exploration_eff}(a) shows the trajectory taken by an agent employing Explore-EQA in a scene in the HM3D dataset taking 30 VLM steps, while ~\Cref{fig:exploration_eff}(b) shows the trajectory taken by an agent employing GraphEQA in the same scene taking a total of 5 VLM steps. We highlight here how Explore-EQA not only takes more steps, but that steps are often guided only by the semantic map constructed by images, leading to more exploration, while GraphEQA takes more structured steps in its environment, guided by the scene graph, to answer the question.

\subsubsection{Choosing object-nodes and frontiers semantically}
In \ref{qual1} we provide qualitative examples of \ourmethod planner choosing to go near a 'couch' to better answer a question about living room lights. This shows that the planner chooses task-relevant objects to guide exploration. In \ref{qual2} we provide a qualitative examples of \ourmethod planner choosing task-relevant frontier nodes and object nodes to guide exploration.

\subsubsection{Qualitative results - Choosing task-relevant object nodes}
\label{qual1}
\begin{minipage}{\textwidth}
\begin{tcolorbox}[mymonobox]

\textcolor{green}{\textbf{Question:}}: Is the living room light on?\\
A. (Do not choose this option)\\
B. (Do not choose this option)\\
C. No\\
D. Yes. \\
 Answer: D \\
\textcolor{blue}{\textbf{At t=0:}}:
\\
\textcolor{red}{\textbf{Agent state}}: The agent is currently at node agent1 at position [-2.928846836090088, -1.5354466438293457, 4.291640281677246]  at room node: room0 with name living room 
\\
\textcolor{red}{\textbf{Action}}: explanation room=I need to further investigate the living room to possibly find indicators like light switches or bulbs.
explanation obj=\textbf{Going near the blue couch might provide a better view of the surroundings, including light fixtures.}
room id=room 0: 'living room' 
object id=object 13: '\textbf{couch}'
\\
\textcolor{red}{\textbf{Answer}}: explanation ans= I need to explore more areas related to lighting in the living room before answering confidently. answer=Answer options A: '(Do not choose this option)' explanation conf='The scene graph and current image do not provide enough information on lighting. Going near specific objects like the couch could help.' confidence level=0.2 is confident=False 
\\
\textcolor{blue}{\textbf{At t=1:}}:\\
\textcolor{red}{\textbf{Agent state}}: The agent is currently at node agent7 at position [-2.2595670223236084, -2.6618852615356445, 4.291640281677246] at room node: room 0 with name living room 
\\
\textcolor{red}{\textbf{Answer}}: explanation ans='The image clearly shows a light fixture on the ceiling that appears to be on. This indicates that the living room light is indeed on.' answer=D 'Yes'
explanation conf='The light fixture being on is visible in the image, providing clear evidence that the living room light is on.' confidence level=0.9
is confident=True
\end{tcolorbox}

\begin{center}
Qualitative example of \ourmethod planner choosing task-relevant object nodes for exploration.
\end{center}
\end{minipage}

\newpage
\subsubsection{Qualitative results - Choosing task-relevant frontier nodes}
\label{qual2}
\begin{minipage}{1.1\textwidth}
\begin{tcolorbox}[mymonobox]
\textcolor{green}{\textbf{Question:}}: Is the kitchen sink currently free of dishes?\\
A. (Do not choose this option)\\
B. (Do not choose this option)\\
C. Yes, it's empty\\
D. No, there are dishes. \\
 Answer: C \\
\textcolor{blue}{\textbf{At t=0:}}:
\\
\textcolor{red}{\textbf{Agent state}}: The agent is currently at node agent 1 at position [0.9115259647369385, -0.7087370157241821, -1.5000251531600952]  at room node: room 0 with name bedroom 
\\
\textcolor{red}{\textbf{Action}}: explanation frontier="The kitchen typically contains objects like chairs, tables, etc. Frontier 4 is near objects that could indicate a transition to a dining or kitchen area, as it is near the couch and chair, which might lead to a kitchen. It's worth exploring this frontier to find the kitchen." frontier id= frontier 4\\
\textcolor{red}{\textbf{Answer}}: explanation ans="Currently, the kitchen is not visible in the scene. The scene graph and current image provide information about a bedroom, but not a kitchen. Therefore, I can't answer whether the kitchen sink is free of dishes." 
answer= A: '(Do not choose this option)' explanation conf="I haven't yet gathered any information on the kitchen or the sink from my current location in the bedroom, and no other regions of the building have been explored yet." 
confidence level=0.0 
is confident=False 

\textcolor{blue}{\textbf{At t=1:}}:\\
\textcolor{red}{\textbf{Agent state}}: The agent is currently at node agent 6 at position [2.5569498538970947, 2.228832721710205, -1.5000251531600952]  at room node: room 0 with name living room 
\\
\textcolor{red}{\textbf{Action}}:
explanation frontier="\textbf{The frontier 2 is connected to objects like the table, cabinet, fridge, and plates, which might lead to a kitchen area or provide a clue to the sink's status.}" frontier id=frontier 2
\\
\textcolor{red}{\textbf{Answer}}:  explanation ans='Currently, there is no visual confirmation of a sink in the image or scene graph, and insufficient detail to determine if the kitchen sink is free of dishes.' 
answer=A: '(Do not choose this option)' explanation conf="The scene graph and current image give no direct view of a kitchen sink or its status, so it's uncertain whether it's free of dishes." confidence level=0.2 
is confident=False

\textcolor{blue}{\textbf{At t=2:}}:\\
\textcolor{red}{\textbf{Agent state}}: The agent is currently at node agent 11 at position [2.6420648097991943, -0.10468387603759766, -1.5000251531600952]  at room node: room 0 with name living room  
\\
\textcolor{red}{\textbf{Action}}:
explanation room='To determine if the kitchen sink is free of dishes, I need to transition from the living room to the kitchen area, where the sink is likely located. Frontiers associated with kitchen-related objects can guide me there.' explanation obj=\textbf{"To check the sink, moving towards the 'plates' near object 41 might lead to the kitchen area."} room id=room 0: 'living room'
object id=object 41: '\textbf{plates}'
\\
\textcolor{red}{\textbf{Answer}}:  explanation ans="Currently, I don't have enough information to directly answer if the kitchen sink is free of dishes, based on the image or scene graph. Further exploration is needed to confirm the state of the sink." answer=A: '(Do not choose this option)'
explanation conf="Given my current location and the items visible, I can't see the sink or its contents. I need to navigate closer to where I suspect the sink might be, based on kitchen-related objects like plates." confidence level=0.3 is confident=False 
\\
\textcolor{blue}{\textbf{At t=3:}}:\\
\textcolor{red}{\textbf{Agent state}}:The agent is currently at node agent 21 at position [1.7573769092559814, -0.35986196994781494, -1.5000251531600952]  at room node: room 0 with name living room  
\\
\textcolor{red}{\textbf{Answer}}:  explanation ans='The image shows the kitchen, and the sink area is visible. There are no dishes in the sink, indicating it is empty.' answer=C: "Yes, it's empty" explanation conf='The image clearly shows the sink, and there are no dishes visible, providing clear evidence to confidently answer the question.'
confidence level=1.0 is confident=True 

\end{tcolorbox}

\begin{center}
Qualitative example of \ourmethod planner choosing task-relevant frontier and object nodes for exploration.
\end{center}
\end{minipage}

\subsection{Additional Discussion}
\para{Novelty of \ourmethod without a learning component:}
Even without a learned component, several aspects of GraphEQA are novel - a real-time, compact, actionable, multimodal memory for grounding VLMs in unseen 3D environments and a semantically-informed efficient exploration strategy. 
Moreover, we argue that modular and easily customizable methodologies like GraphEQA are effective for zero-shot deployment in novel real-world environments without requiring expensive end-to-end training, as reflected in our real-world experiments.
The performance of certain modules can be improved with task-specific fine-tuning, however, that can come at the cost of open-world generalizability.

\para{How can \ourmethod handle multiple questions in the same episode?} To handle multiple questions, our method can be extended by maintaining a separate visual memory per question and discarding as each question is answered. However, handling temporally extended questions that require an unknown number of images remains an important direction for future work.

\para{Utility of using both binary and numeric confidence:}
Using both binary and numeric confidence for episode termination is empirically motivated. In practice, this redundancy improved the reliability of numerical scores and provided a more robust indicator of the VLM's confidence.

\para{Broader applicability of \ourmethod beyond EQA tasks:}
Although we mainly focus on EQA tasks, GraphEQA is broadly applicable to other navigation tasks such as object-goal navigation, leveraging its ability to perform grounded exploration in unseen 3D environments. In particular, EQA presents a greater challenge than object-goal navigation, as it requires more complex semantic and spatial reasoning, including identification, counting, existence, state, and functional understanding (see OpenEQA [9], Fig. 5). 
Compared to GraphEQA, methods like ConceptGraphs \cite{gu2023conceptgraphsopenvocabulary3dscene}, which rely on \textit{enriched} scene graphs without visual context, face key limitations on EQA tasks: 1) Enriching scene graphs with relational information via LLMs may suffice for object-goal navigation but falls short for EQA, which often demands nuanced, fine-grained semantic understanding that may not be captured in the graph. While task-specific enrichment is possible, it is not scalable across diverse EQA tasks.
GraphEQA addresses this challenge by maintaining semantically-rich task-relevant images in its multi-modal memory and using a VLM for in-context reasoning. 2) Open-vocabulary 3D instance segmentation, used in ConceptGraphs, is computationally expensive -- even when applied incrementally -- and limits the real-time applicability of these methods. GraphEQA maintains a light-weight yet semantically rich multi-modal memory which can be constructed in real-time.

\vspace{-10pt}
\section{Real-world Experiments in Home Environments}
\vspace{-4pt}
\label{app:questions}
Experiments for Home Environments (a, b) and the questions asked of GraphEQA are given below. We provide a sequence of ten images from the head camera on the robot to illustrate exploration and validation of GraphEQA's answer to the question. Each experiment was executed twice successfully. We show a single trial from this set for each experiment. For videos of real-world experiments, please refer to the \href{https://saumyasaxena.github.io/grapheqa/}{website}.
\vspace{-8pt}
\subsection{Home Environment (a)}
\begin{tcolorbox}
[colback=white!5!white,colframe=black!75!black, coltitle=white, fonttitle=\bfseries, rounded corners, boxrule=1pt, boxsep=1pt]
\textbf{Question 1:} \textit{Is there a blue pan on the stove?} \\
A. Yes\\
B. No \\
 Answer: \textbf{A}
\end{tcolorbox}
\begin{figure*}[h]
    \centering
\includegraphics[width=\textwidth]{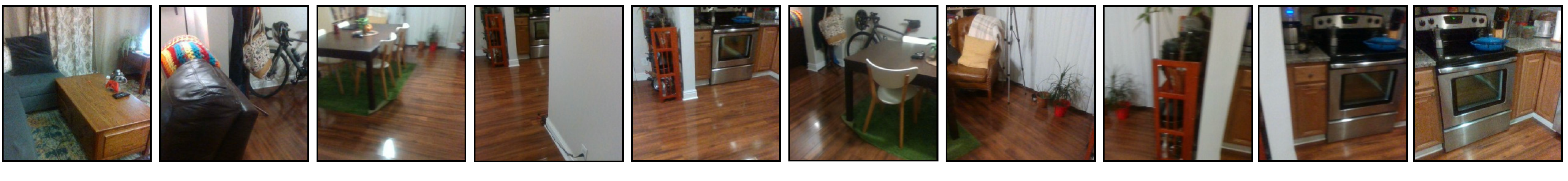}
\vspace{-15pt}
    \caption{\footnotesize{\textbf{Home Environment (a) Question 1: Is there a blue pan on the stove?} For this question, the agent takes four VLM steps in the environment, beginning by  reasoning about the current information it has access to; ``I can't answer confidently as the current view and scene graph don't reveal a kitchen or stove.''. The agent then takes two GotoFrontierNodeStep steps to explore, and as shown in Frame 5 (from the left) gains access to the stove via the scene graph: ``The stove is connected to region and frontier nodes, suggesting proximity.''. The agent then takes a GotoObjectNodeStep(stove) step, navigates to the stove, and upon reaching it answers \textbf{``The image shows a blue pan on the stove.''}, answering with high confidence.}}
    \label{fig:homea_bluepan}
\end{figure*}

\begin{tcolorbox}
[colback=white!5!white,colframe=black!75!black, coltitle=white, fonttitle=\bfseries, rounded corners, boxrule=1pt, boxsep=1pt]
\textbf{Question 2:} \textit{How many white cushions are there on the grey couch?} \\
A. 1 \\
B. 2 \\
C. 3 \\
D. 4 \\
 Answer: \textbf{B}
\end{tcolorbox}
\begin{figure*}[h]
    \centering
\includegraphics[width=\textwidth]{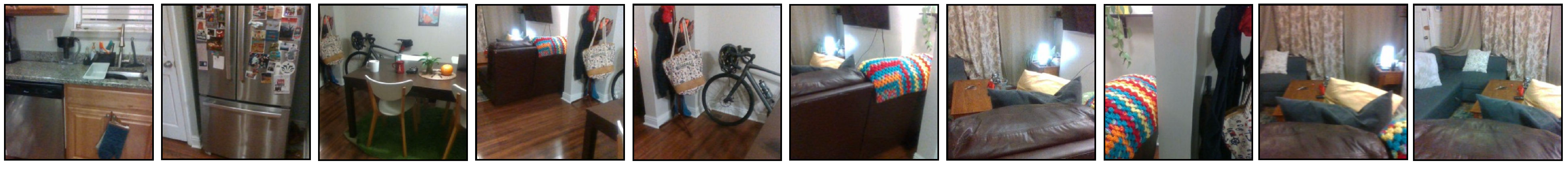}
\vspace{-15pt}
    \caption{\footnotesize{\textbf{Home Environment (a) Question 2: How many white cushions are there on the grey couch?} For this question, the agent takes a total of six VLM steps, initiating its exploration by reasoning about the frontiers and objects it has access to in the scene graph; ``No frontiers or objects are directly related to locating the grey couch. I'll choose a frontier to explore unexplored areas.''. The agent then has access to a couch (see Frame 4 from the left) and chooses GotoObjectNodeStep(couch) for two VLM steps to search around the only couch it can see. After executing these object node steps, the agent then has access to a secondary couch it has not explored, and so chooses GotoObjectNodeStep(couch) once more to explore the grey couch, stating ``The current view doesn't clearly show the number of white cushions on the grey couch, so I need a closer look.'' In the final VLM step the agent answers the question; \textbf{``The image shows a grey couch with two white cushions. There is also a table with various items in front of the couch.''}}}
    \label{fig:homea_cushions}
\end{figure*}

\begin{tcolorbox}
[colback=white!5!white,colframe=black!75!black, coltitle=white, fonttitle=\bfseries, rounded corners, boxrule=1pt, boxsep=1pt]
\textbf{Question 3:} \textit{Where is my handbag?} \\
A. On the coat rack \\
B. On the floor \\
C. On the couch \\
 Answer: \textbf{A}
\end{tcolorbox}
\begin{figure*}[h]
    \centering
\includegraphics[width=\textwidth]{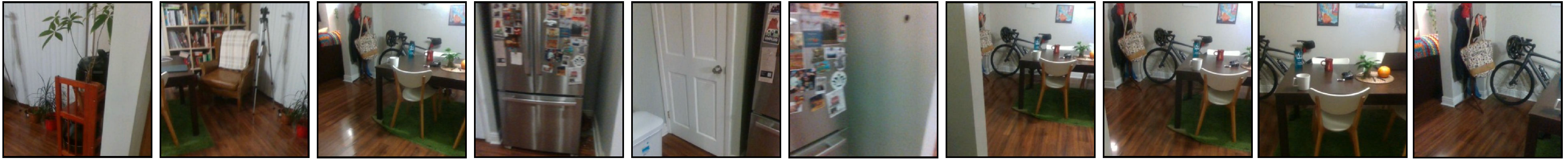}
\vspace{-15pt}
    \caption{\footnotesize{\textbf{Home Environment (a) Question 3: Where is my handbag?} The agent takes a total of three VLM steps to determine the location of the handbag, beginning with exploration guided by the scene graph; ``The scene graph shows several frontiers connected to objects or areas, but none seem directly related to a potential couch or coat rack. Exploration of frontiers is necessary''. After the first GotoFrontierNodeStep to explore, the agent identifies a handbag in the environment; ``Although the scene graph indicates the handbag is in the living room, further exploration is needed to confirm its position.'', but cannot yet confirm its position among the options available to it. Finally, the agent takes a GotoObjectNodeStep to gain a better view of the handbag and its position; \textbf{``I have visual confirmation from the current image showing the handbag on the coat rack.''. }To see the full experiment in video, please see \href{https://saumyasaxena.github.io/grapheqa/}{website}.}}
    \label{fig:homea_handbag}
\end{figure*}
\newpage
\begin{tcolorbox}
[colback=white!5!white,colframe=black!75!black, coltitle=white, fonttitle=\bfseries, rounded corners, boxrule=1pt, boxsep=1pt]
\textbf{Question 4:} \textit{Where is the trashcan?} \\
A. Next to the sink \\
B. Next to the refrigerator \\
 Answer: \textbf{A}
\end{tcolorbox}
\begin{figure*}[h]
    \centering
\includegraphics[width=\textwidth]{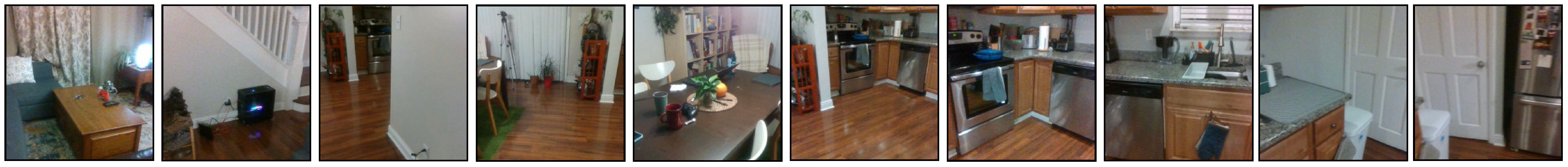}
\vspace{-15pt}
    \caption{\footnotesize{\textbf{Home Environment (a) Question 4: Where is the trashcan?} The agent takes a total of six VLM steps to determine the location of the trashcan, beginning with exploration guided by the scene graph; ``The image and scene graph don't provide information about a kitchen or a trashcan location. Choosing a frontier that might lead to a kitchen.'' The following four steps are GotoFrontierNodeStep actions, as the agent prioritizes exploring over investigating specific objects in the environment; ``The current scene graph shows objects like a cushion, stove, chair, table, blanket, and other kitchen-related items like a dishwasher, hand towel, cabinet, and sink in the vicinity.'' Finally, the agent visually confirms the location of the trashcan; \textbf{``Given the clear visibility of the trashcan next to the sink in Image 1, I am confident in answering with certainty.''} Note that \textit{Image 1} for this experiment is the 9th image in the sequence of ten above.}}
    \label{fig:homea_trashcan}
\end{figure*}

\begin{tcolorbox}
[colback=white!5!white,colframe=black!75!black, coltitle=white, fonttitle=\bfseries, rounded corners, boxrule=1pt, boxsep=1pt]
\textbf{Question 5:} \textit{Is the front door next to the staircase open?} \\
A. Yes \\
B. No \\
 Answer: \textbf{B}
\end{tcolorbox}
\vspace{-8pt}
\begin{figure*}[h]
    \centering
\includegraphics[width=\textwidth]{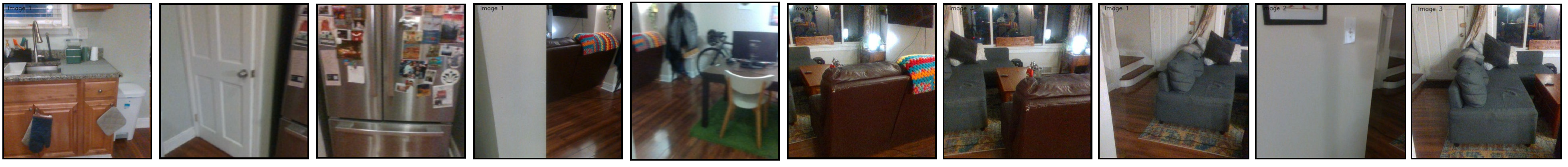}
\vspace{-15pt}
    \caption{\footnotesize{\textbf{Home Environment (a) }Question 5:\textbf{ Is the front door next to the staircase open?} The agent takes a total of four VLM steps to determine the state of the door, beginning its exploration by identifying where it is located and choosing an action appropriately; ``The scene graph shows the agent's location in the kitchen. There aren't any nodes indicating a door or staircase.  I will explore a frontier as it could lead to a new area where the front door or staircase might be visible''. The agent's next two actions are GotoFrontierNodeStep actions, explaining the second of these choices as follows; ``To locate the staircase, I will explore the frontier connected to a couch, which could be in the living room area.''. The agent then chooses another frontier since ``...the current view does not show much that relates directly to the front door or staircase.'' Along its trajectory to that frontier node, the agent identifies relevant images to the question and answers; \textbf{``The current image shows a blue couch in the center. To the left, a staircase is visible. Beyond the couch is a door that is closed.''}.}}
    \label{fig:homea_door}
\end{figure*}

\newpage
\subsection{Home Environment (b)}
\begin{tcolorbox}
[colback=white!5!white,colframe=black!75!black, coltitle=white, fonttitle=\bfseries, rounded corners, boxrule=1pt, boxsep=1pt]
\textbf{Question 1:} \textit{What is the color of the dehumidifier?} \\
A. Blue \\
B. White and Gray \\
 Answer: \textbf{B}
\end{tcolorbox}
\begin{figure*}[h]
    \centering
\includegraphics[width=\textwidth]{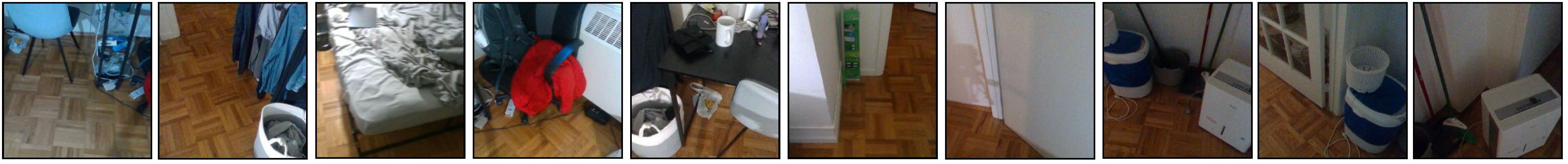}
\vspace{-15pt}
    \caption{\footnotesize{\textbf{Home Environment (b) Question 1: What is the color of the dehumidifier?} The agent takes a total of nine VLM steps to answer this question, and starts by exploring a frontier node; ``I need to find the dehumidifier machine to determine its color. It is not currently in the scene graph or visible.'' The next eight steps are chosen as frontier steps, with explanations like ``The dehumidifier is not in the scene graph. I need to explore to find it. The current image shows a potential candidate object (white appliance) near the chair (object3) and backpack (object17).'' The final frontier step taken reveals the dehumidifier; ``The object potentially representing the dehumidifier (object26, labeled 'box') is in room0. There are two white, boxy machines on the floor, likely dehumidifiers or air purifiers. One has a gray top panel.''}}
    \label{fig:homeb_dehumidifier}
\end{figure*}
\begin{tcolorbox}
[colback=white!5!white,colframe=black!75!black, coltitle=white, fonttitle=\bfseries, rounded corners, boxrule=1pt, boxsep=1pt]
\textbf{Question 2:} \textit{Is there any object on the black plastic drawer?} \\
A. Yes \\
B. No \\
 Answer: \textbf{B}
\end{tcolorbox}
\begin{figure*}[h]
    \centering
\includegraphics[width=\textwidth]{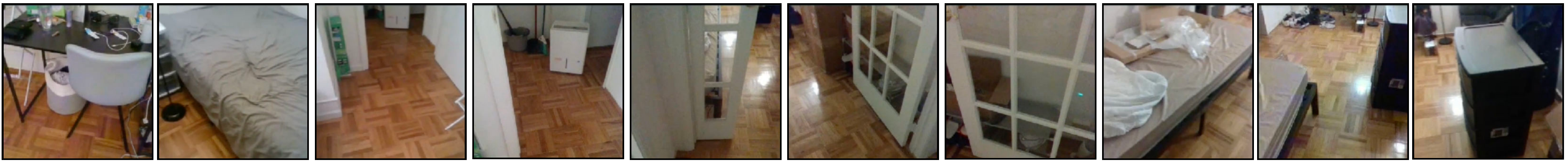}
\vspace{-15pt}
    \caption{\footnotesize{\textbf{Home Environment (b) Question 2: Is there any object on the black plastic drawer?} The agent takes a total of nine VLM steps. The first step taken is a frontier node step chosen due to its connection to a candidate  object; ``I need to explore the environment to find the 'black plastic drawer' mentioned in the question, as it is not currently in the scene graph. The image shows a potential candidate object near the 'chair' (object1). Frontier3 is linked to the 'chair' (object1) and appears to be in the direction of the area needing investigation based on the current view.'' The following six steps are GotoFrontierNodeStep actions with explanations similar to ``This object is not currently in the scene graph or visible. Exploring a frontier might reveal new areas and objects, potentially including the drawer.'' Finally, the agent is able to visually confirm the presence of the black plastic drawer with high confidence. }}
    \label{fig:homeb_plastic_drawer}
\end{figure*}
\begin{tcolorbox}
[colback=white!5!white,colframe=black!75!black, coltitle=white, fonttitle=\bfseries, rounded corners, boxrule=1pt, boxsep=1pt]
\textbf{Question 3:} \textit{What is next to the white shopping bag?} \\
A. Stool \\
B. Broom \\
C. Dehumidifier \\
 Answer: \textbf{A}
\end{tcolorbox}
\begin{figure*}[h]
    \centering
\includegraphics[width=\textwidth]{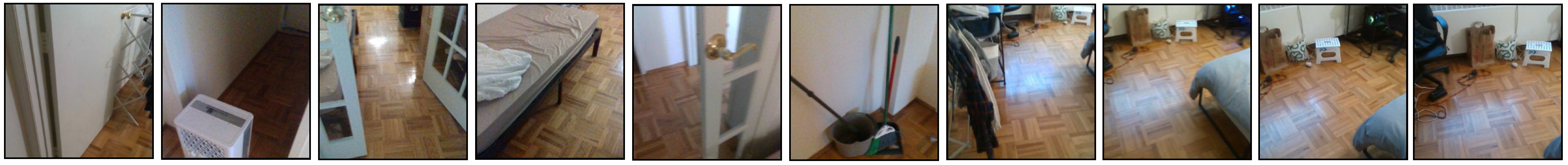}
\vspace{-15pt}
    \caption{\textbf{Home Environment (b) Question 3: What is next to the white shopping bag?} The agent takes a total of four VLM steps in this experiment, all of which are exploratory GotoFrontierNodeStep actions until the stool is identified as being next to the shopping bag.}
    \label{fig:homeb_whitebag}
\end{figure*}

\newpage
\section{Additional Real-world Experiments in Office and Workshop Environments}
\label{app:supprealworldeqa}
Additional experiments in the workshop and office environment are shown in \Cref{fig:workshop_env} and \Cref{fig:office_env} respectively. They include both success and failure cases.

\subsection{Workshop environment}
\begin{figure}[h]
    \centering
\includegraphics[width=1.0\textwidth]{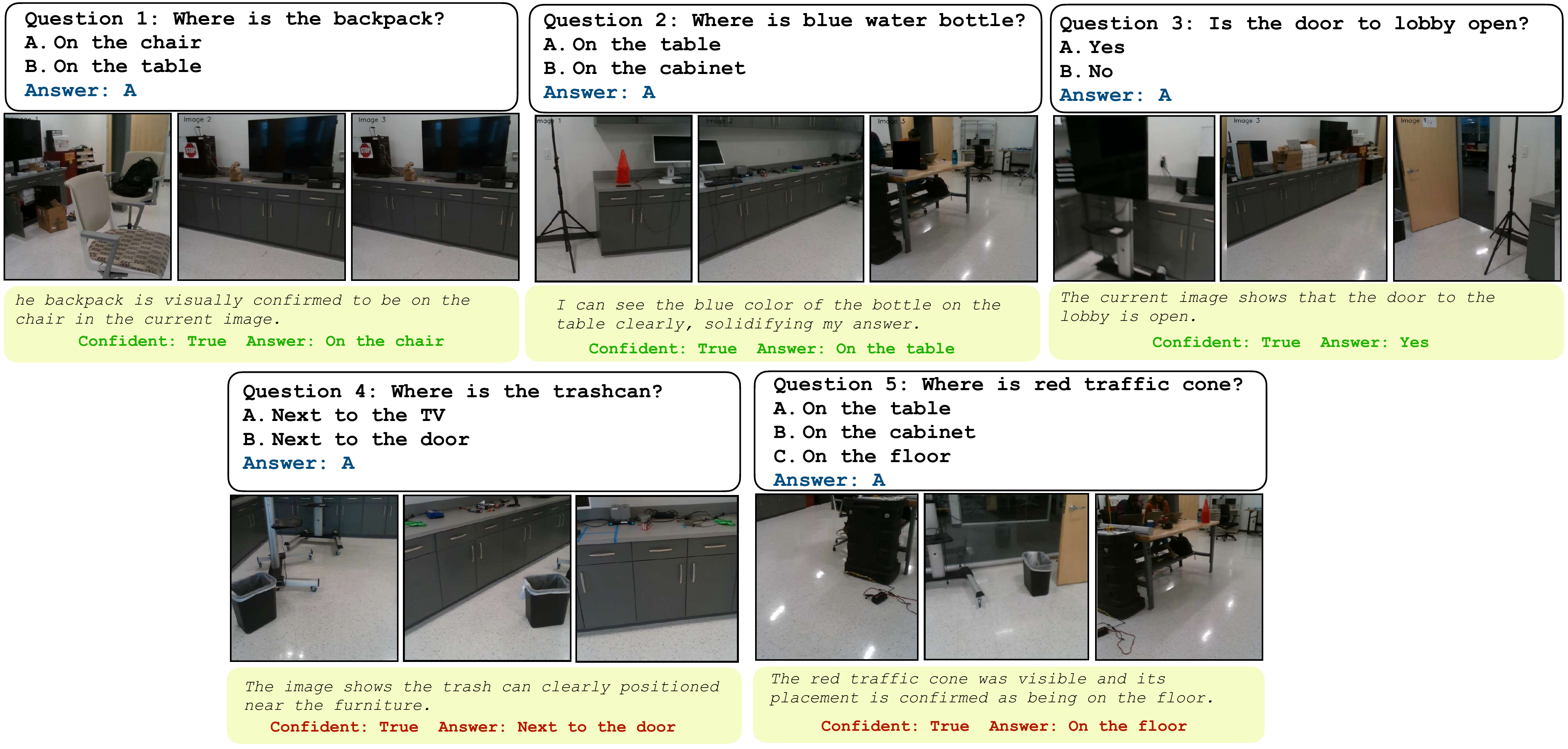}
    \caption{\footnotesize{\textbf{Workshop environment:} Each set of images is from the head camera on the Stretch robot, and represents the task-relevant images at the last planning step. Provided under the images are the answers, confidence levels, and explanations output from the VLM planner.}}
    \label{fig:workshop_env}
\end{figure}

\begin{tcolorbox}
[colback=white!5!white,colframe=black!75!black, coltitle=white, fonttitle=\bfseries, rounded corners, boxrule=1pt, boxsep=1pt]
Question 1: \textit{Where is the backpack?} \\
A. On the chair \\
B. On the table \\
 Answer: \textbf{A}
\end{tcolorbox}
The relevant images for the task are shown in \Cref{fig:workshop_env}. The agent first takes a {\small\verb|<Goto_Object_node>(chair)|} step in the environment after an initial rotate-in-place mapping operation to populate the scene graph, choosing to investigate a chair found in the environment. An explanation for this choice is provided by the VLM: ``Objects like the chair or table might have the backpack, so checking close to these areas is essential.'' The robot then begins navigating to the chair to determine if the backpack is located there. During execution of the trajectory toward the chair, \ourmethod leverages its task-relevant visual memory to score three images encountered on its way to the chair. 
After finishing the execution of this trajectory, \ourmethod answers the question with `On the chair' and provides the following explanation of its answer: ``The backpack is visually confirmed to be on the chair in the current image. The presence of the backpack on the chair makes it clear that the correct answer to the question is `On the chair'.'' 


\begin{tcolorbox}
[colback=white!5!white,colframe=black!75!black, coltitle=white, fonttitle=\bfseries, rounded corners, boxrule=1pt, boxsep=1pt]
Question 2: \textit{Where is the blue water bottle?} \\
A. On the table\\
B. On the cabinet\\
C. On the floor\\
Answer: \textbf{B} 
\end{tcolorbox}

As shown in \Cref{fig:workshop_env}, after exploring the environment with one {\small\verb|<Goto_Object_node>(table)|} step, the agent successfully finds the water bottle and confirms its location, providing the following justification for its answer: ``The image shows a table with some objects on top, including a blue water bottle. There is also a computer monitor and various tools visible on a countertop''. Qualitative results for additional questions are provided in \Cref{fig:workshop_env}. 

\subsection{Office environment}

\begin{tcolorbox}
[colback=white!5!white,colframe=black!75!black, coltitle=white, fonttitle=\bfseries, rounded corners, boxrule=1pt, boxsep=1pt]
Question 1: \textit{Is my sweater on the blue couch?}\\
A. Yes\\
B. No\\
Answer: \textbf{A}
\end{tcolorbox}
The agent starts by taking a {\small\verb|<Goto_Object_node>(couch)|} step, to explore the blue couch. The following VLM explanation of the object step clarifies \ourmethod is referring to the blue couch: \texttt{explanation\_obj=`I need to locate the blue couch before I can determine if the sweater is on it or not.' object\_id=<object\_node\_list.object\_1: couch>}

\begin{figure}[h]
    \centering
\includegraphics[width=1.0\textwidth]{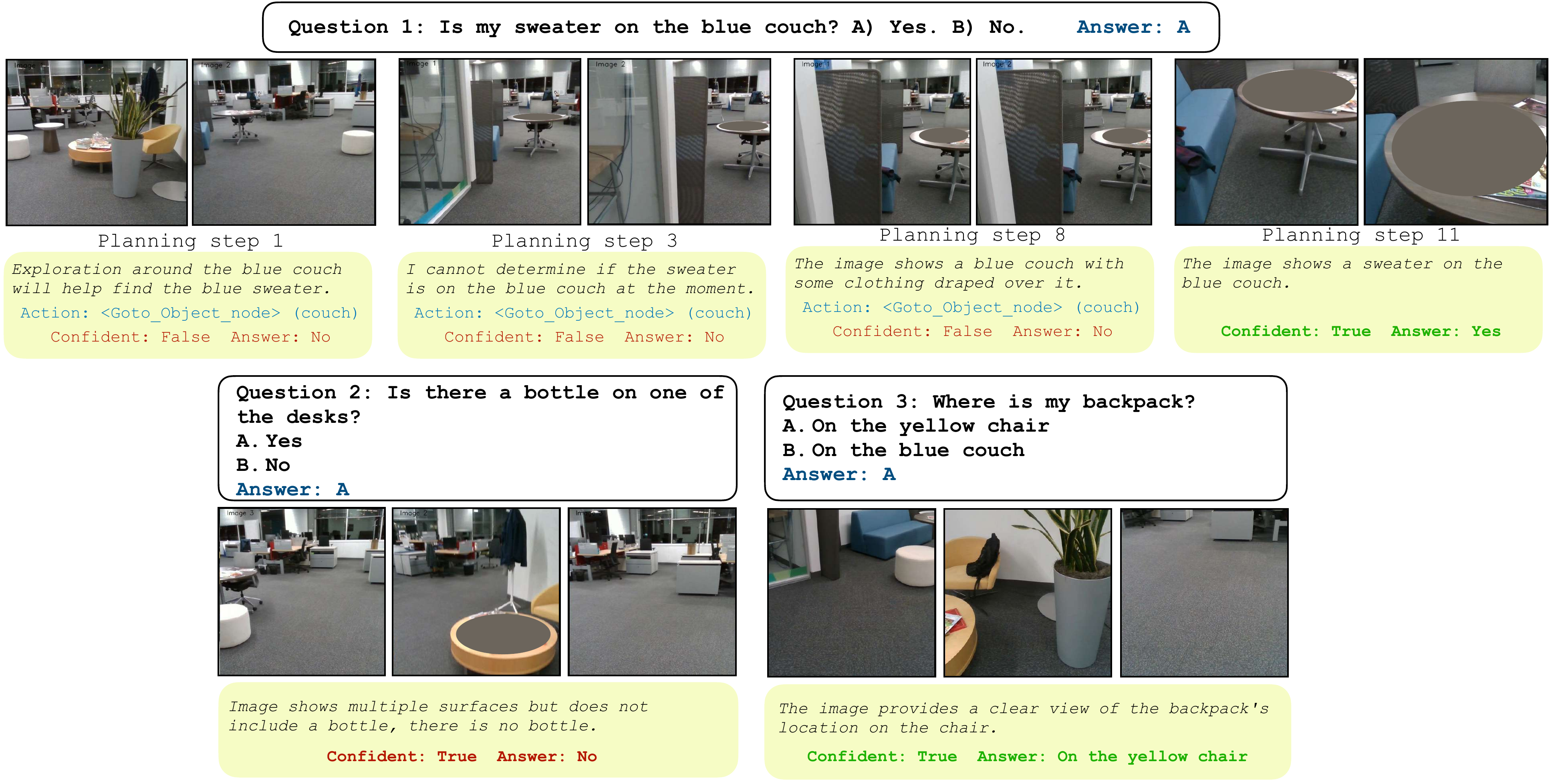}
    \caption{\footnotesize{\textbf{Office environment:} Each set of images for Question 1 is from the head camera on the Stretch robot, and represents the top-k task-relevant images at each planning step. For questions 2 and 3, the images correspond to the relevant images at the last planning step. Provided under the images are the answers, confidence levels, and explanations output from the VLM planner.}}
    \label{fig:office_env}
\end{figure}

The low-level planner implementation on Hello Robot's stretch does not plan the entire path to the blue couch, however, resulting in several more {\small\verb|<Goto_Object_node>(couch)|} steps before answering the question confidently after 11 steps. Qualitative results for additional questions are provided in \Cref{fig:office_env}.

\end{document}